\documentclass[10pt,twocolumn,letterpaper]{article}
\pdfoutput=1
\usepackage{authblk}

\usepackage{iccv}
\usepackage{times}
\usepackage{epsfig}
\usepackage{graphicx}
\usepackage{amsmath}
\usepackage{amssymb}

\usepackage{pifont}
\usepackage{multirow}
\usepackage{rotating}

\usepackage{caption}
\usepackage{subcaption}

\renewcommand{\arraystretch}{1}

\usepackage[pagebackref=true,breaklinks=true,letterpaper=true,colorlinks,bookmarks=false]{hyperref}
\hypersetup{pdfauthor={Tuan-Hung Vu, Anton Osokin, Ivan Laptev},pdftitle={Tube-CNN: Modeling temporal evolution of appearance\\ for object detection in video}}

\iccvfinalcopy 

\begin{document}
	
	\title{Tube-CNN: Modeling temporal evolution of appearance\\ for object detection in video}
	\author{Tuan-Hung Vu\thanks{Currently at valeo.ai, France. This work was done while T-H was at INRIA.} \qquad  Anton Osokin\thanks{Currently at NRU HSE, Russia. This work was done while Anton was at INRIA.} \qquad  Ivan Laptev\vspace{.2cm}\\
		INRIA/ENS, Paris, France\vspace{-.0cm}\\}
	\maketitle
	\thispagestyle{empty}
	
	\begin{abstract}
		Object detection in video is crucial for many applications.
		Compared to images, video provides additional cues which can help to disambiguate the detection problem.
		Our goal in this paper is to learn discriminative models for the temporal evolution of object appearance and to use such models for object detection.
		To model temporal evolution, we introduce space-time tubes corresponding to temporal sequences of bounding boxes.
		We propose two CNN architectures for generating and classifying tubes, respectively.
		Our tube proposal network (TPN) first generates a large number of spatio-temporal tube proposals maximizing object recall.
		The Tube-CNN then implements a tube-level object detector in the video.
		Our method improves state of the art on two large-scale datasets for object detection in video: HollywoodHeads and ImageNet VID.
		Tube models show particular advantages in difficult dynamic scenes.
	\end{abstract}
	
	\vspace{-0cm}
	\section{Introduction}
	\label{sec:intro}
	\vspace{-0cm}
	Object detection has been recently advanced by the large progress of convolutional neural networks (CNNs).
	In the last few years, the performance of object detectors has doubled on some benchmarks~\cite{girshick2014rcnn,girshick2015fast,gidaris2015locnet} and the technology became mature for deployment in real applications such as automatic car driving.
	Despite this success, accurate object detection remains a difficult challenge in situations with partial occlusions, unusual viewpoints, motion blur and difficult lighting.

	\begin{figure}[t!]
		\centering
		\includegraphics[trim = 88mm 80mm 110mm 93mm, clip, width=\linewidth]{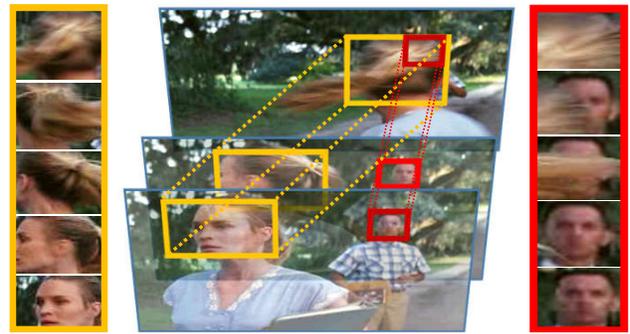}\vspace{.1cm}
		\caption{Space-time object tubes used in our work to capture evolution of object appearance over short sequences of video frames.}
		\vspace{-.cm}
		\label{fig:teaser}
	\end{figure}
	
	Static images may sometimes lack information to identify the presence of an object.
	Video, on the other hand, provides additional cues which can help to disambiguate the detection problem.
	For example, sequences with objects undergoing dynamic occlusions and rotations are structured and contain space-time patterns with object-specific properties, see Figure~\ref{fig:teaser}.
	In this paper, we argue that dynamic changes of appearance provide discriminative information that can boost object detection in difficult situations.
	
	Object tracking in video can compensate some failures of object detectors by imposing temporal smoothness.
	Tracking, however, often assumes the temporal continuity of appearance and can be misleading when such assumptions break due to strong occlusions or fast motions.
	In contrast to tracking, our goal in this paper is to learn discriminative models for the {\em temporal evolution of object appearance} and to use such models for object detection.
	
	To model the temporal evolution of appearance, we introduce space-time {\em tubes} corresponding to sequences of object bounding boxes in consecutive frames.
	We propose two CNN architectures for generating and classifying tubes, respectively.
	Our tube proposal network (TPN) first generates a large number of spatio-temporal tube proposals maximizing the objectness score and guaranteeing high recall for the ground truth object tubes.
	The Tube-CNN implements a tube-level object detector in the video using tube proposals as input.
	Examples of tubes are shown in Figure~\ref{fig:teaser}.

	People are among the most frequent and difficult objects in images and video.
	Changes in clothing, poses and hair-style as well as frequent occlusions and dynamic variations of appearance make person detection particularly challenging.
	In this work, we pay close attention to this object class.
	To span the variety of dynamic scenes and activities, we consider people in Hollywood movies and address detection of person heads.
	To train our models we use the recent HollywoodHeads dataset~\cite{vu2015context} and extend it with the annotation of head tracks in $3.8$ hours of video from $21$ different movies.
	We test our method on two datasets and report significant improvements compared to the state of the art.
	We also evaluate and demonstrate advantages of our method in difficult scenes with strong dynamic occlusions.
	
	In addition, the new dataset from ImageNet challenge of object detection in video (VID) \cite{russakovsky2015videoImageNet} allows us to train and test tube models on $30$ other object classes.
	The dataset contains short video snippets fully annotated with object bounding-box tracklets.
	Our tube models show notable improvement in comparison with frame-level detector baselines.
	We achieve comparable results with the winner of ImageNet VID 2015 challenge~\cite{kang2016tcnn}.
	On the Youtube-Objects (YTO) dataset~\cite{prest2012learning}, a weakly annotated dataset closely related to ImageNet VID dataset, our framework surpasses state-of-the-art localization performance by a large margin.
	
	\vspace{-0cm}
	\section{Related works}
	\label{sec:related_work}
	\vspace{-0cm}
	Object detection in still images is one of the most active subjects in computer vision.
	Since the success of R-CNN~\cite{girshick2014rcnn}, many works have advanced the region-based CNN model.
	SPP-net~\cite{he2014sppnet} and Fast R-CNN~\cite{girshick2015fast} speed up R-CNN by a large magnitude with the idea of performing regional-pooling on a feature map.
	Ren \textit{et al.}~\cite{ren2015faster} propose the Region Proposal Network (RPN) removing the need for ad-hoc bottom-up object proposals, and increasing the speed of object detectors. DeepMask~\cite{DeepMask} and SharpMask~\cite{SharpMask} are examples of methods aiming to generate class-agnostic segmentation mask proposals.
	Although using segmentation proposals could be beneficial, those methods heavily depend on large scale segmentation datasets like COCO~\cite{lin2014microsoft} to achieve satisfactory result.
	Motivated by previous works, we propose two models that can generate and classify spatio-temporal tube proposals for object detection in video.
	
	Object detection in video is closely related to object detection in images.
	One can directly use object detector on every single frames.
	Video provides additional cues, such as temporal continuity.
	A common practice for video object detection is to post-process detections produced by object detector on single image with trackers~\cite{Bojanowski_2013_ICCV,ordonez2011im2text,everingham2006hello, pirsiavash2011globally}.
	The tracker first links high-confident detections on consecutive frames into tracks. As object detector often produces unstable scores along an object track, the linking step helps finding good detections by filling in the gaps.
	An additional smoothing step could be done to stabilize object position on the track.
	Recently, Kang \textit{et al.}~\cite{kang2016object} use this practice to generate object tracks. 
	In their work, the tracker output is named \textit{tube proposal}.
	It then becomes input for the next post-processing steps including box-perturbation, max pooling and rescoring.
	To avoid possible confusion, we stress that the term \textit{tube proposal} is used in our work with a different meaning. 
	Unlike other video detection methods~~\cite{pirsiavash2011globally,tang2015subgraph}, which separate appearance and motion processing phases, our models treat video signal as an entire combination of the two factors.
	
	Although there are many works on object proposals in still images (such as SelectiveSearch~\cite{uijlings2013selective}), choice of methods for video tube proposals is limited.
	Most of existing tube proposal methods~\cite{oneata:hal-01021902,gemert2015apt} focus on action localization rather than object detection.
	The parallel work of Kang \textit{et al.}~\cite{kang2017object} is probably the closest to us.
	In this work, the authors propose a Tubelet Proposal Networks for generating tubelet proposals, which are then scored by an encoder-decoder CNN-LSTM.
	Regression is done on every frame to adjust box location along the tubelet.
	The authors show that their proposal network performs best with temporal window size of $5$ frames.
	In our work, we argue that in such short temporal intervals, a linear movement assumption is good enough to produce well-localized object tube proposals for detection.
	Our best spatio-temporal tube proposals linearly span on $10$ temporal frames.
	Although the proposal network in~\cite{kang2017object} share the same spirit with our TPN model, our model is much simpler while being efficient.
	Compared to~\cite{kang2017object}, we show significant improvements on ImageNet VID and Youtube-Objects datasets.
	
	Object detection in video and \emph{tracking-by-detection} have a tight relation.
	Graph-based methods~\cite{pirsiavash2011globally,tang2015subgraph} formulate object tracking as a global optimization problem on association graph.
	The association graph is constructed by measuring appearance similarity between objects.
	In~\cite{leal2016learning}, the authors propose a Siamese CNN to learn such appearance similarity efficiently.
	In this work, we consider tracking-by-detection methods as useful tools to improve object detection performance.
	We focus on object detection and do not aim at improving tracking performance.
	Our model design and parameter choices are in favor of this goal.
	
	\vspace{-0cm}
	\section{Tube-CNN for object detection}
	\label{sec:tubemodel}
	\begin{figure*}[t]
		\begin{center}
			\includegraphics[trim = 0cm 0cm 0cm 0cm, clip, width=0.95\textwidth]{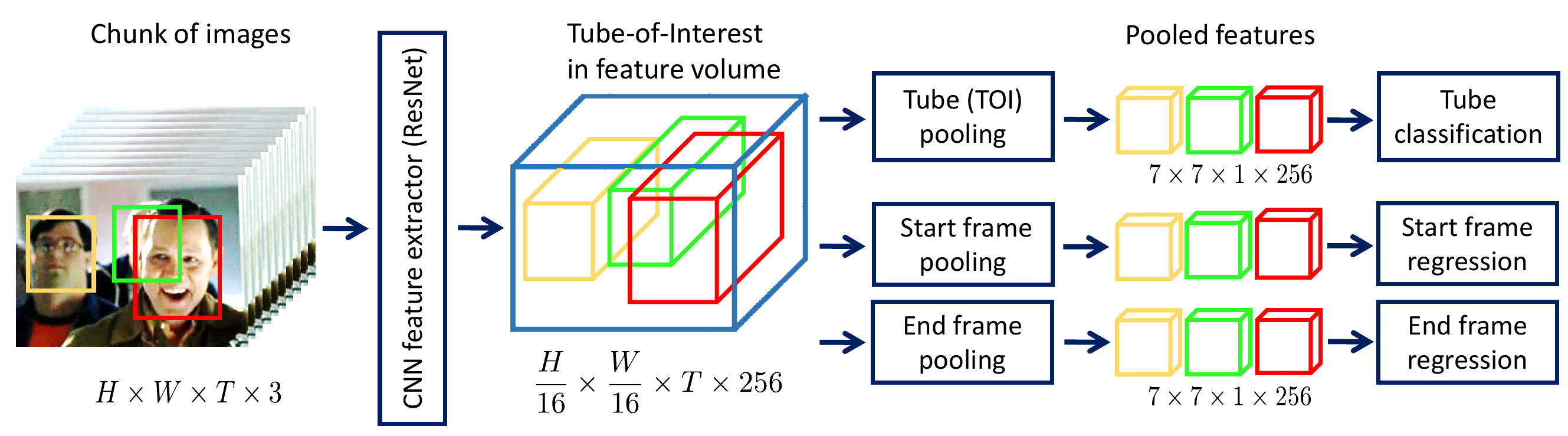}
		\end{center}
		\caption{Architecture of Tube-CNN for object detection.
			The input is a chunk of images and a set of tube proposals.
			The model starts with the CNN feature extractor to get a spatio-temporal feature volume.
			Afterwards the model splits into the three branches: tube classification (starts with Tube-of-Interest, TOI, pooling), regression on the start and end frames, which begins with the frame-level Region-of-Interest pooling, ROI.}
		\label{fig:tubecnn_archi}
	\end{figure*}
	
	We now describe the Tube-CNN model for detecting objects in video.
	Our model operates on short chunks of consecutive video frames.
	Instead of rectangular image regions, our elementary units are spatio-temporal tubes spanning several frames.
	In this section, we assume that a set of candidate tubes (\emph{tube proposals}) is provided and describe our Tube-CNN that classifies and refines them.
	
	The number of all possible tubes is huge and does not allow to consider them exhaustively.
	In this paper, we consider only linear tubes, i.e., tubes corresponding to uniform linear motion in the video.
	Such an approximation is reasonable only locally for small chunks of consecutive frames.
	In what follows, we always represent a linear tube by the two rectangles on its first and last frames.
	We study the effect of different tube lengths in Section~\ref{sec:discuss}.
	
	\subsection{Architecture}
	\label{sec:tube_cnn}
	Tube-CNN is a convolutional neural network operating on chunks of consecutive frames and the corresponding sets of tube proposals.
	For each proposal, the model outputs the class scores and the refined proposal position.
	
	Tube-CNN is an end-to-end model consisting of three main blocks: CNN feature extraction, tube classification and tube regression.
	The overall network architecture is shown in Figure~\ref{fig:tubecnn_archi}.
	
	\vspace{-0.35cm}\paragraph{CNN feature extractor.}
	The first block of the network extracts a feature map independently for every frame in the input chunk.
	Extracted features are stacked along the temporal dimension to form a spatio-temporal feature volume.
	To achieve good performance, we reuse the idea of sharing computations from first CNN layers among all proposals~\cite{he2014sppnet,girshick2015fast}.
	For the architecture on this block, we can reuse most of single frame CNNs (we try CaffeNet\footnote{\url{https://github.com/BVLC/caffe/tree/master/models/bvlc_reference_caffenet}} and ResNet~\cite{He2015resnet}).
	
	\paragraph{Tube classification.}
	Classification starts with Tube-of-Interest (TOI) pooling on the feature volume.
	Within the TOI-pooling layer, we first map the coordinates of tube proposals to subvolumes of features.
	Next, we max-pool within each frame to obtain a fixed-size feature map.
	The last stage consists in the temporal aggregation of the maps coming from all the frames.
	We have tried max-pooling, average pooling, and 1-dimensional convolution.
	Our conclusion is that temporal max-pooling works best (without significant difference although) and we use it in what follows.
	Note, that if temporal aggregation is done by max or average pooling, the number of parameters in Tube-CNN does not depend on the chunk length~$T$, and the learned model can be readily applied to chunks of arbitrary length.
	
	After TOI-pooling, we obtain a feature map of a fixed size for each tube proposal.
	This map is fed into the final part of the classification network.
	For example, in the case of ResNet-101 pipeline~\cite{He2015resnet}, it is the 4-th block.
	The tube classification branch ends with the cross-entropy loss.
	
	\paragraph{Tube regression.}Tube regression adjusts spatial positions of tube proposals to better localize the object.
	Tube regression consists of two networks for bounding box regression~\cite{girshick2015fast} at the two ends of the tube.
	Both branches of tube regression start with ROI pooling~\cite{girshick2015fast} on the corresponding frame (the start and end frames of the chunk).
	Tube regression branches end with the smooth L1 loss~\cite{girshick2015fast}.
	
	\subsection{Supervision}
	\label{sec:tube_cnn_train}
	To train the Tube-CNN model, one needs a dataset with annotated object tracks.
	Details of the datasets used for our training are given in Section~\ref{sec:dataset}.
	
	At the training time, Tube-CNN model takes a chunk of $T$ consecutive frames and a set of tube proposals as inputs.
	Given the chunk, we find a set of ground-truth tracks passing through it.
	Each ground-truth track is approximated within the chunk by a linear ground-truth tube.
	We use these tubes to assign labels to tube proposals in this chunk.
	
	Label assignment for a tube proposal is based on the tube overlap between the tube proposal and the ground-truth tubes.
	We define the \emph{tube overlap} between two tubes as minimum of the spatial Intersection-over-Union (IoU) overlaps at their ends.
	Each sampled tube proposal of a training chunk is assigned with a label: positive (class label of an object) and negative (background).
	Tube proposals having tube overlap $\theta_{GT} \ge 0.5$ with the best matched ground-truth are marked as positives.
	Tube proposals with $0.1 \le \theta_{GT} < 0.5$ are marked as negatives.
	Additional details on the training setup are given in Section~\ref{sec:train_detail} and in the Appendix~\ref{sec:train_tube_cnn}.
	
	\section{Generating tube proposals}
	The Tube-CNN model uses tube proposals as input.
	Tube proposals are defined as the sequences of region proposals on consecutive video frames that hypothesize spatio-temporal positions of objects.
	To cut down the overall number of possible proposals, we consider only the ones corresponding to uniform linear motion.
	
	In this section, we introduce a CNN model that generates tube proposals given an input chunk of frames: Tube Proposal Network.
	We also propose a baseline tracking-based approach to build tube proposals on top of single-frame region proposals.
	Evaluation of different tube proposal methods is reported in Section~\ref{sec:recall_ana}.
	
	\subsection{Tube proposal network}
	\label{sec:tpn}
	We now present our Tube Proposal Network (TPN).
	This is a CNN model that can generate a small set of high quality class-agnostic tube proposals used for object detection.
	
	\label{sec:TPN_archi}
	\begin{figure}[t]
		\hspace*{-0.5cm}
		\begin{center}
			\includegraphics[trim = 0cm 0cm 0cm 0cm, clip, width=\columnwidth]{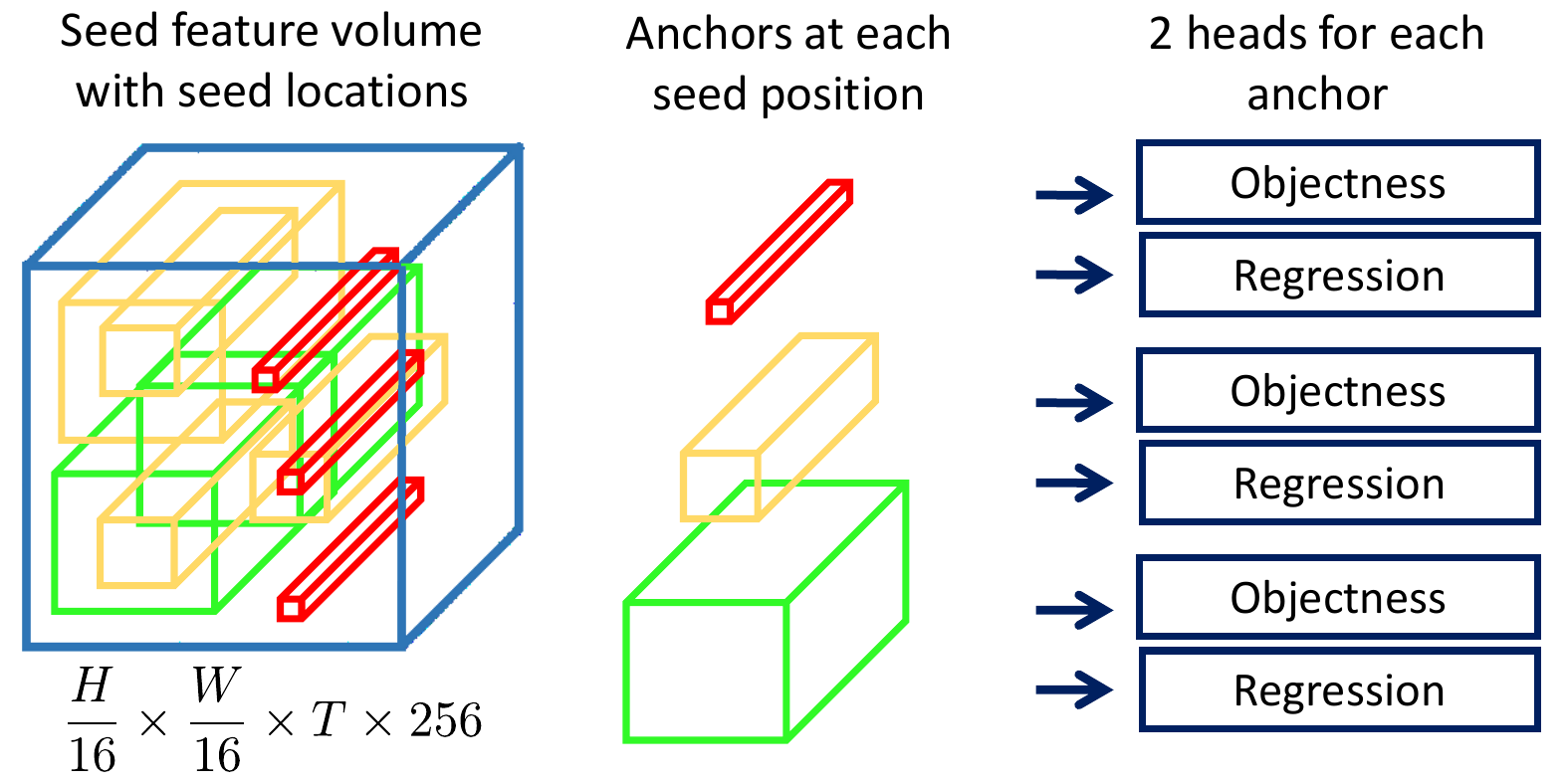}
		\end{center}
		\caption{
			Architecture of Tube Proposal Network (TPN).
			Tube anchors are plotted in red, yellow and green.
		}
		\label{fig:tpn_archi}
	\end{figure}
	
	\paragraph{Architecture.}
	Tube Proposal Network is a fully convolutional network~\cite{long2015fully}, taking a chunk of consecutive frames as input and producing a set of tube proposals.
	Each tube proposal has an objectness score and a refined region location.
	Figure~\ref{fig:tpn_archi} illustrates the TPN model.
	
	Similar to the Tube-CNN model, TPN passes input chunks of consecutive frames through several CNN layers.
	The output feature maps are stacked along the temporal dimension and passed through a $3 \times 3 \times 3$ volumetric convolutional layer (conv3D)~\cite{tran2015c3d} to form a feature volume.
	
	We refer to all spatial positions of the feature volume as \emph{seed locations} and the corresponding features as \emph{seed features}.
	Each seed feature vector is associated with a set of $K$~reference tubes, which we call \emph{tube anchors}.
	Tube anchors sharing the same center axis in the feature volume correspond to the same seed location as well as the same seed feature vector.
	
	Each seed feature vector is then passed through an \emph{anchor network} to produce two outputs per each of the $K$~tube anchors: the objectness score and parameters for the tube regression.
	In our setting, the anchor network consists of two fully-connected layers.
	Finally, we obtain the tube proposals by collecting the regressed locations of all the tube anchors.
	Scores of the tube proposals directly come from the objectness scores of the anchors.
	
	To prune tube proposals, we apply Non Maximum Suppression (NMS) based on the objectness score.
	We modify the standard NMS to the case of tubes by replacing the spatial IoU overlap ratio with the tube overlap ratio.
	We provide more details in the Appendix~\ref{sec:train_tpn}..
	
	\paragraph{Training.}
	TPN and Tube-CNN have many common properties, thus, their training procedures are similar.
	Compared to Tube-CNN operating with tube proposals, TPN is trained on mini-batches of tube anchors.
	
\begin{figure}[t]
		\hspace*{-0.5cm}
		\begin{center}
			\includegraphics[trim = 75mm 85mm 75mm 85mm, clip, width=\columnwidth]{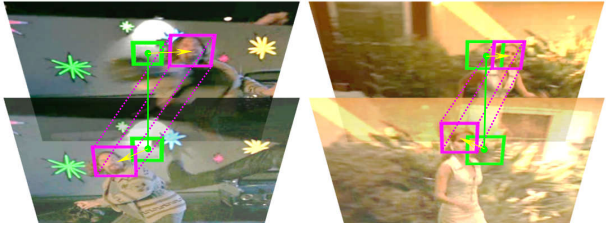}
		\end{center}
		\caption{Examples of TPN proposal.
			Examples of TPN proposals. Tube anchors and final proposal are plotted by green and magenta respectively.}
		\label{fig:tpn_example}
	\end{figure}

	Each seed feature vector is connected to $K$ tube anchors.
	Tube anchors have the same center axis with the receptive field of the seed features.
	In the related work~\cite{ren2015faster}, anchors for region proposals correspond to several scales and aspect ratios.
	Adopting the setup to tubes, we have tried to add anchors with motion between the start and end regions, but it has not improved the recall while significantly increasing the number of anchors, thus complexity at the testing stage.
	We therefore use tube anchors, with spatial location fixed, but with varying scales and aspect ratios.
	Figure~\ref{fig:tpn_example} shows examples of initial tube anchors and the corresponding final TPN proposals.
	We show that TPN can regress an anchor without motion to a tube proposal with motion, because the receptive field of the anchor is typically much bigger than the anchor itself and, thus, TPN can get enough information about the spatio-temporal neighborhood.
	
	To train, we label each tube anchor as either positive (object) or negative (background).
	The label assignment procedure is similar to one of the Tube-CNN training procedure.
	The only difference is that the supervision of TPN is class-agnostic.
	Positive anchors have tube overlap ratio $\theta_{GT} \ge 0.5$ with a ground-truth tube of any object class.
	Anchors having $\theta_{GT}\le 0.3$ with all ground-truth tubes are labeled as negatives.
	
	\subsection{Tube proposals by tracking box proposals}
	\label{sec:tube_from_track_box}
	
	We introduce a tracking-based method to produce tube proposals from region proposals (which we refer to as \emph{box proposals}) of single frames.
	Given a chunk of $T$ consecutive frames, we generate box proposals of the start frame by the Selective Search method~\cite{uijlings2013selective}.
	For each of those box proposals, we hypothesize its corresponding positions on the end frame by using a tracker.
	Specifically, we use the Kanade-Lucas-Tomasi (KLT) tracker~\cite{shi1994good} to obtain point tracks between the start and the end frames of the chunk.
	Each box proposal is associated with a set of interior point tracks, corresponding to a set of movement directions.
	Those directions are then clustered into $N_b$ directional bins.
	We apply RANSAC on $N_h$ most populated bins to construct $N_h$ candidate temporal paths in the interval of $T$ frames\footnote{We use $N_b=16$ and $N_h=4$ in our experiments}\!\!.
	Tube proposals are then constructed by linearly moving the box along all hypothesized temporal paths.
	
	The above process could be built upon the output of any box proposal method.
	To have a rough comparison with~\cite{kang2016tcnn}, we generate tube proposals from high-scoring detections produced by our Fast R-CNN baseline\footnote{This baseline is described in details in Section~\ref{sec:det_res}}\!\!.
	
	\section{Experiments}
	\label{sec:experiment}
	In this section, we present and analyze experimental results of our method.
	We first introduce video datasets for object detection in Section~\ref{sec:dataset}.
	We then clarify our experimental setup in Section~\ref{sec:train_detail}.
	Section~\ref{sec:recall_ana} evaluates the quality of tube proposals.
	Section~\ref{sec:det_res} presents our main results for object detection and compares them to the state of the art.
	In Section~\ref{sec:discuss} we present an ablation study analyzing parameters and design choices of our method.

	\subsection{Datasets}
	\label{sec:dataset}
	\paragraph{HollywoodHeads dataset.}
	HollywoodHeads\footnote{\scriptsize\url{http://www.di.ens.fr/willow/research/headdetection/}}~\cite{vu2015context} is a recent large-scale dataset for person head detection with video clips from 21 movies.
	The original dataset provides head annotation for all the frames.
	To train tube models in this work, we extend the annotation by combining detections to form tracks.
	Our extended HollywoodHeads dataset contains $331\;746$ video frames with complete annotation of head tracks, split into $216\;719$,\,\,\,$67\;181$ and $47\;846$ frames for training, validation and test respectively.
	For evaluation and comparison to the previous work, we keep the test set of~\cite{vu2015context} with $1\;302$ annotated frames\footnote{We have removed $10$ test frames from the original test set, because they belonged to short video clips of length less than $10$ frames.}\!\!.
	
	Occlusions are among the most challenging factors for object detection.
	To study the detection performance under occlusions, we select a subset of video clips, named HollywoodHeads-Hard, with partially visible heads.
	HollywoodHeads-Hard is automatically composed from clips with multiple annotated heads, where ground-truth bounding boxes have significant overlap on the horizontal axis.
	In total, we obtain $266$ difficult clips and select one frame for testing from each of them.
	
	\paragraph{Casablanca dataset.} In addition to HollywoodHeads, we evaluate head detection on the Casablanca dataset. This dataset was introduced in~\cite{ren2008finding} and later extended in~\cite{vu2015context} with the annotation of missing ground-truth heads.
	
	\paragraph{ImageNet VID dataset.} We also evaluate our method on the multi-class object detection benchmark.
	ImageNet VID~\cite{russakovsky2015videoImageNet} is a large-scale dataset for object detection providing complete annotation of 30 object classes in $1.3$ million video frames.
	The dataset contains $5\;354$ short videos split into $3\;862$ videos for training, $555$ videos for validation and $937$ videos for testing.
	The $30$ object classes of the ImageNet VID dataset are a subset of $200$ classes used in ImageNet DET challenge.
	As the ground-truth for the test set is not published, we evaluate and compare our results on the validation set with $176\;126$ frames.
	
	\paragraph{Youtube-Objects dataset.} 
	Youtube-Objects (YTO) is a video dataset for object localization providing annotation for $10$ object classes on a sparse set of key frames~\cite{prest2012learning}.
	The $10$ object classes in YTO form a subset of 30 object classes in ImageNet VID.
	We use YTO for additional evaluation of the models pretrained on ImageNet VID.
	
	\subsection{Training details}
	\label{sec:train_detail}
	\paragraph{Base networks.}
	In our experiments, Tube-CNN is initialized either from CaffeNet~\cite{girshick2015fast} or from Resnet-101~\cite{He2015resnet} models pre-trained on ImageNet~\cite{imagenet_cvpr09}. The TPN model is based on the CaffeNet architecture.
	
	\paragraph{Training parameters.}
	We optimize Tube-CNN and TPN with the stochastic gradient descent (SGD) algorithm with momentum 0.9 and weight decay 0.0005 on mini-batches~\cite{krizhevsky2012imagenet}.
	The classification and regression are trained with the cross-entropy loss and smooth L1 loss~\cite{girshick2015fast}, respectively.
	We fix tube length $T=10$ for all tube models.
	This design choice is analyzed in Section~\ref{sec:discuss}.
	
	\paragraph{Hard negative mining.} Hard negative mining enables training on difficult samples and has been proven effective for object detection with HOG and DPM models~\cite{dalal2005histograms,girshick2012discriminatively} and more recently with CNN methods~\cite{arandjelovic2015netvlad,shrivastava2016training}.
	In our experiments, we find that a few iterations of hard negative mining improves detection performance for all of our models.
	More details are provided in the Appendix~\ref{sec:hardnegdetails}.
	Section~\ref{sec:discuss} analyzes the effect of hard negative mining on the detection performance.
	
	\subsection{Evaluation of tube proposals}
	\label{sec:recall_ana}
	In this section, we evaluate the quality of tube proposals produced by different methods on HollywoodHeads and ImageNet VID datasets.
	
	\paragraph{Tube proposal baselines.}
	Tube proposals have been previously proposed for the task of action detection in video.
	We here evaluate and compare two state-of-the-art methods for action proposals~\cite{gemert2015apt,oneata:hal-01021902}.
	For the 3D-Proposal method~\cite{oneata:hal-01021902}, we extract $10,000$ tube proposals for every shots of $10$ frames.
	For APT~\cite{gemert2015apt}, we use longer shots due to the length constraint imposed by the method.

	\paragraph{Evaluation of recall.}
	Object proposals should maximize the coverage of ground-truth object bounding boxes.
	We evaluate the quality of tube proposals by examining their recall on both frame level and video level.
	We refer to the two measures as \textit{box-recall} and \textit{tube-recall}, respectively.
	Tube-recall is the percentage of ground-truth tubes having tube overlap-ratio more than $0.5$ with at least one tube proposal (tube overlap-ratio is defined in Section~\ref{sec:tube_cnn_train}).
	To compute box-recall, we first split tube proposals into per-frame box proposals.
	Box-recall is then computed as the percentage of still images ground-truth bounding boxes having IoU overlap-ratio more than $0.5$ with at least one box proposal.
	\begin{figure}[t]
		\hspace*{-0.5cm}
		\begin{tabular}{ccc}	
			&\hspace{0.8cm}\textit{\small HollywoodHeads} & \hspace{0.2cm}\textit{\small ImageNet VID}\\
			\raisebox{1.15cm}{\rotatebox[origin=c]{90}{\footnotesize{Box-recall}}}&\includegraphics[trim = 0cm 0cm 0cm 0cm, clip, width=0.23\textwidth]{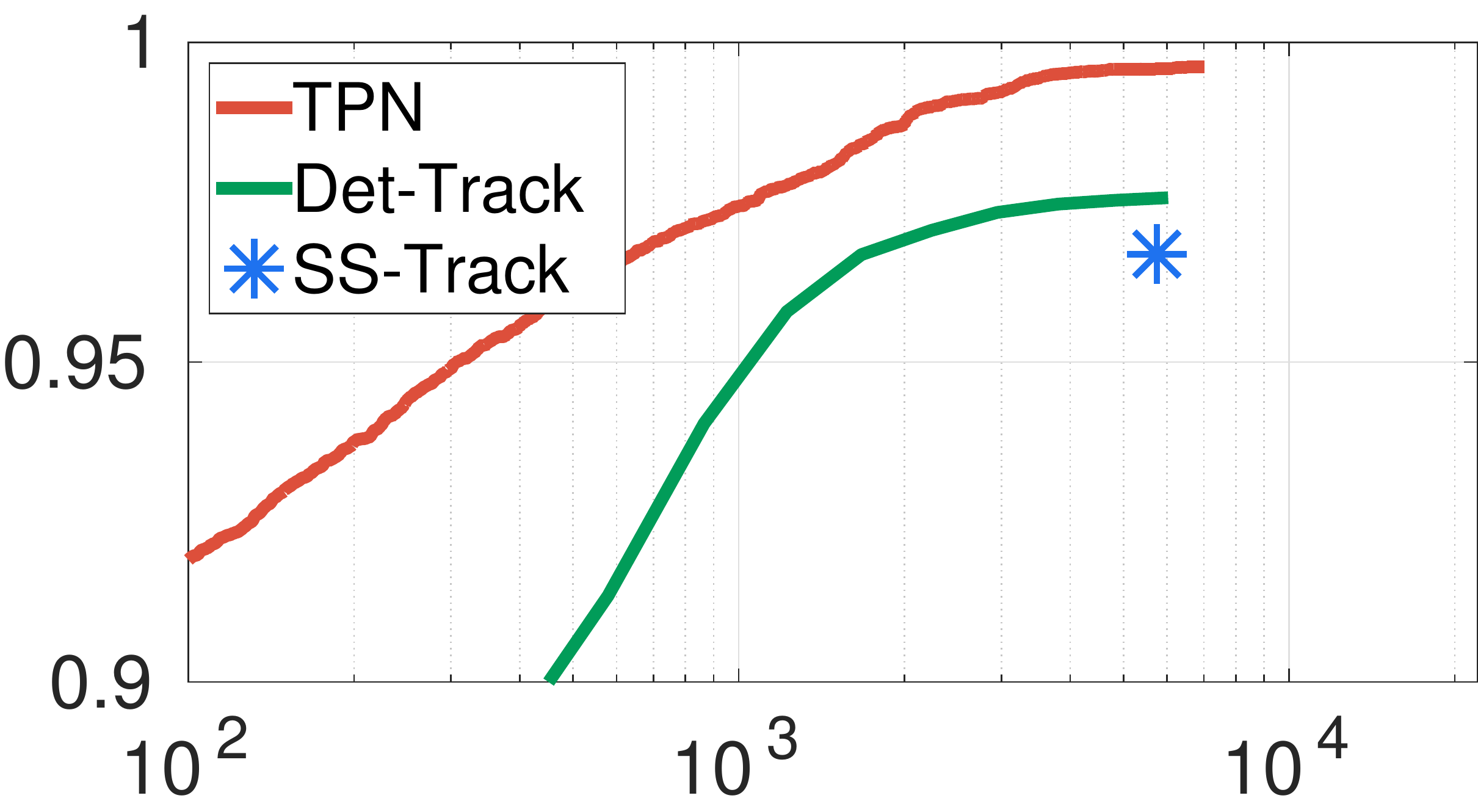}&\hspace{-0.3cm}\includegraphics[trim = 0cm 0cm 0cm 0cm, clip, width=0.23\textwidth]{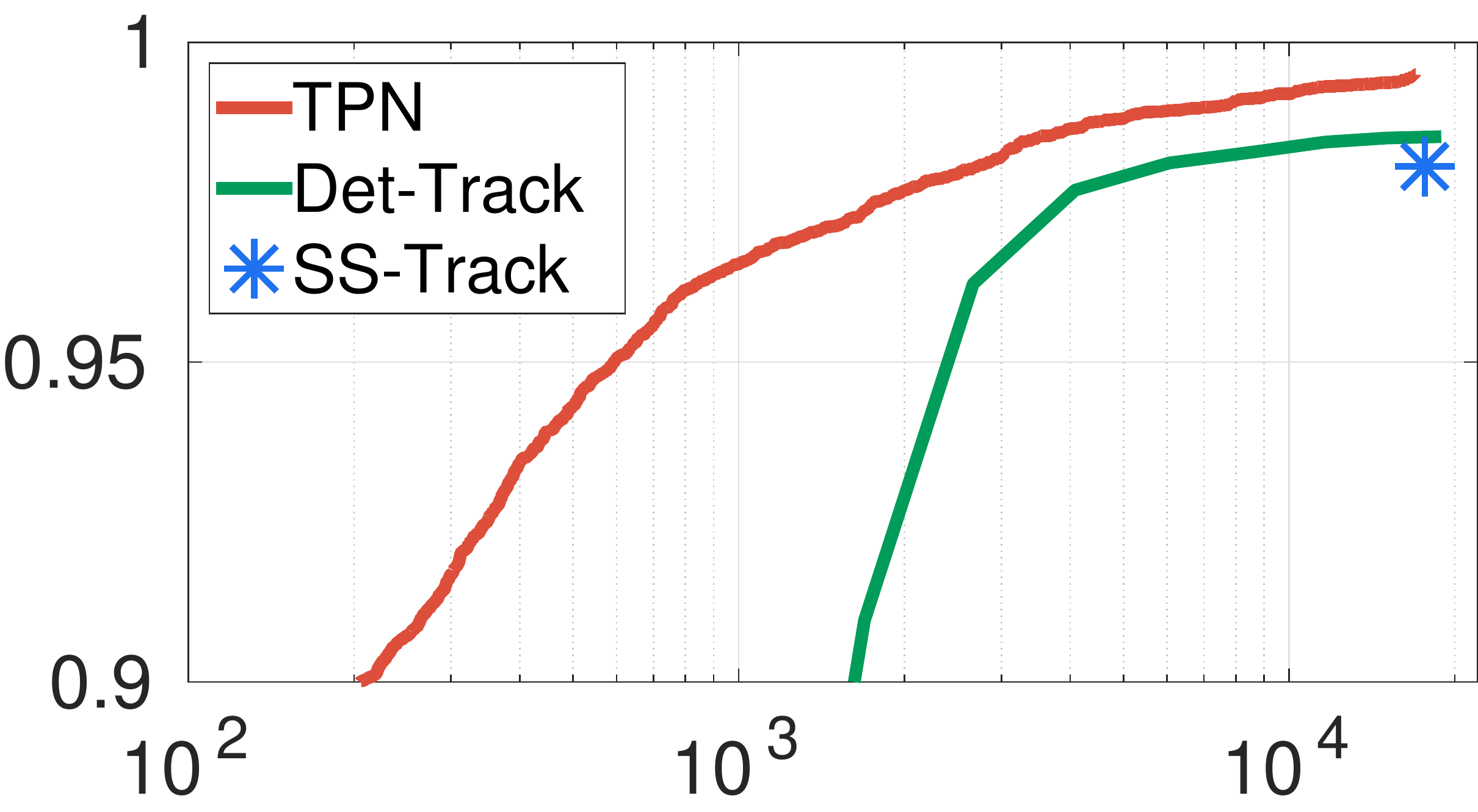}\\
			\raisebox{1.15cm}{\rotatebox[origin=c]{90}{\footnotesize{Tube-recall}}}&\includegraphics[trim = 0cm 0cm 0cm 0cm, clip, width=0.23\textwidth]{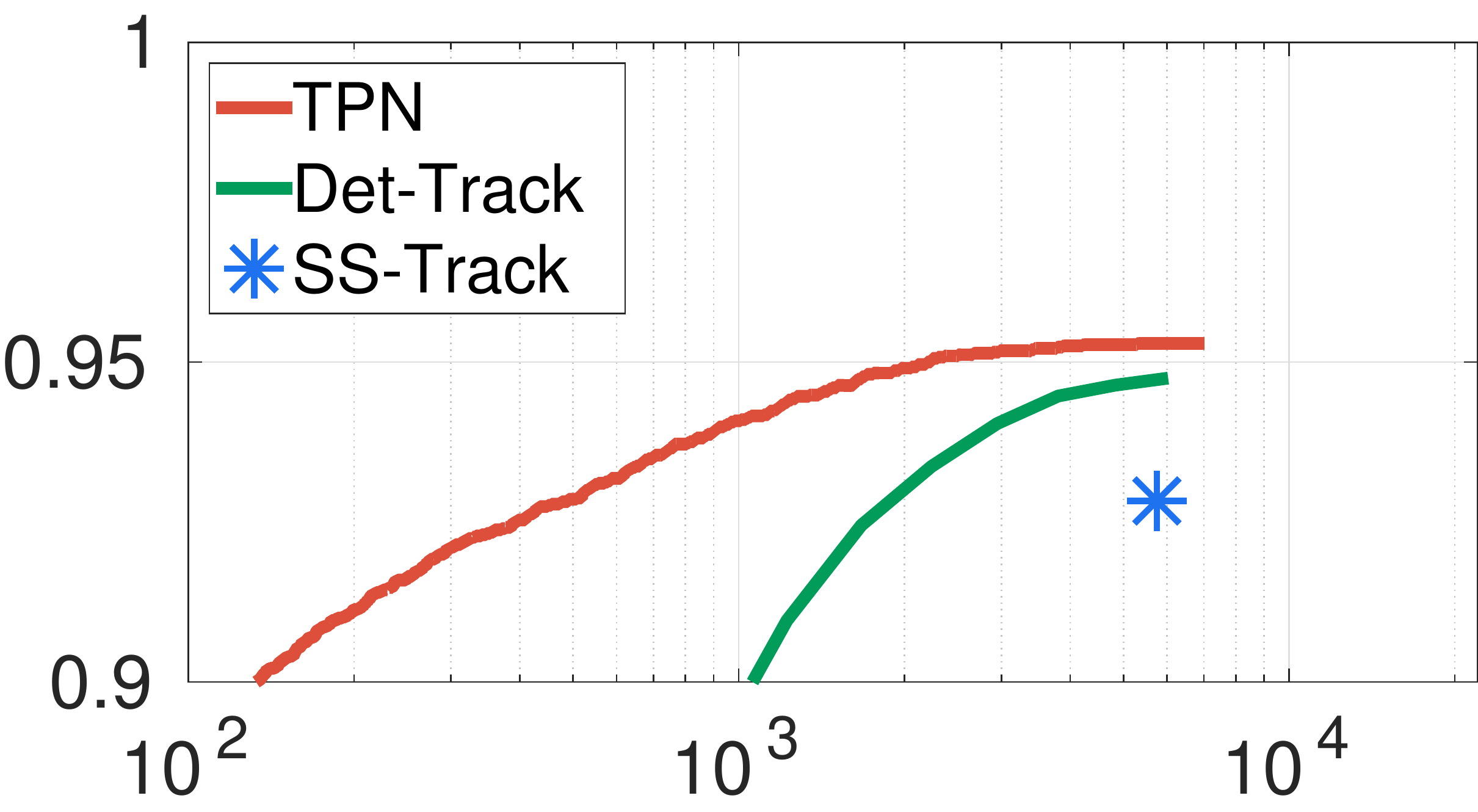}&\hspace{-0.3cm}\includegraphics[trim = 0cm 0cm 0cm 0cm, clip, width=0.23\textwidth]{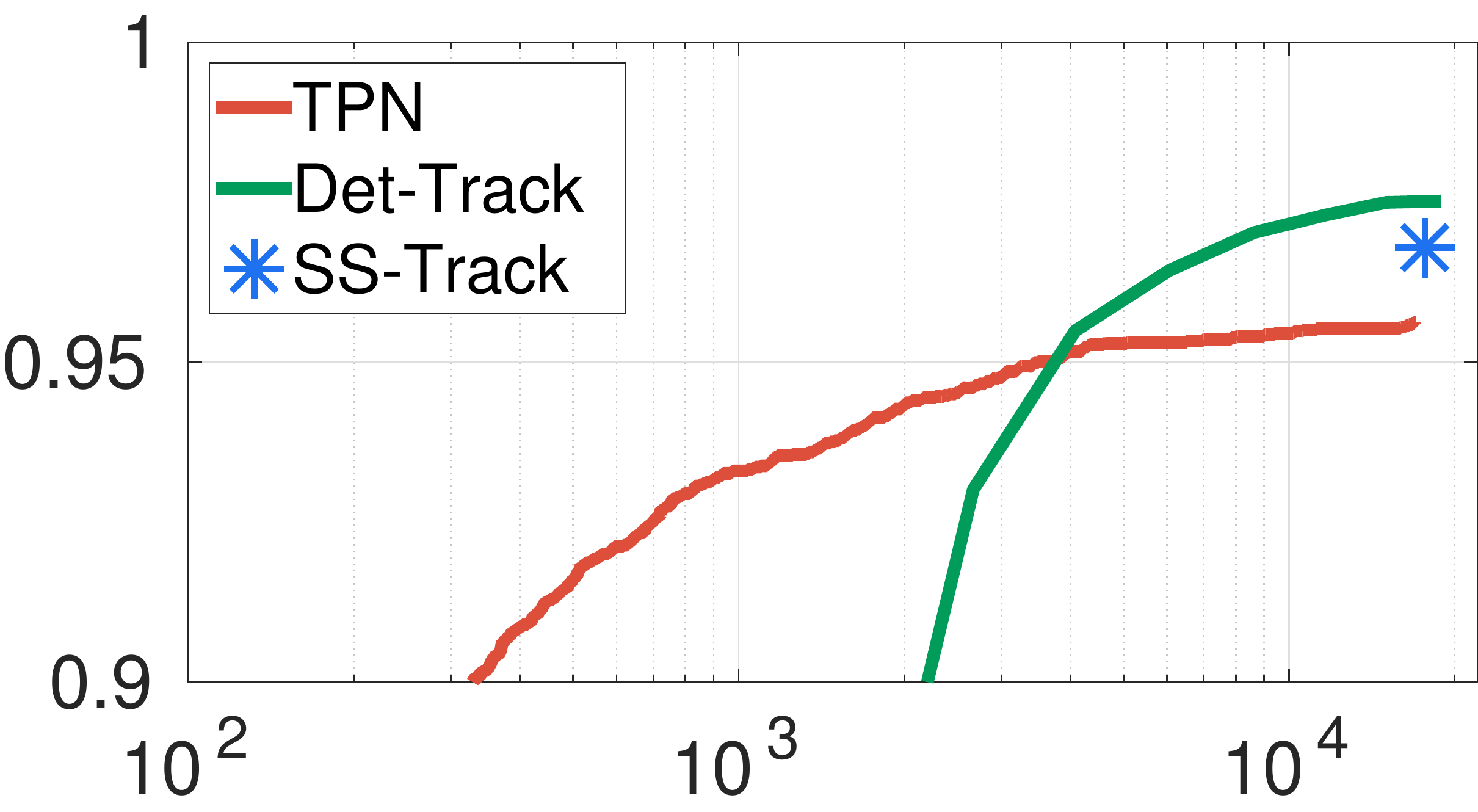}\\
			&\hspace{0.6cm}\footnotesize{Number of proposals} & \hspace{0.3cm}\footnotesize{Number of proposals}\\[-0.0cm]
		\end{tabular}
		\caption{Box-recall and tube-recall \textit{vs.} average number of tube proposals per chunk on HollywoodHeads and ImageNET VID datasets.}
		\label{fig:recall_ana}
	\end{figure}
	
	\begin{table}[t]
		\begin{center}
			\begin{tabular}{c||r|cc}
				\parbox{2.6cm}{\centering Method} & Avg.\# TP & Tube Rec. & Box Rec.\\
				\hline
				\hline
				3D Proposals~\cite{oneata:hal-01021902} & 10K & 55.3 & 61.3 \\
				APT Proposals~\cite{gemert2015apt} & 4.2K & 48.7 & 62.1 \\
				\hline
				SS-Track (ours) & 6.7K & 92.8 & 96.7 \\
				Det-Track (ours) & 6K & 94.8 & 97.6  \\
				\hline
				\multirow{4}{*}{TPN (ours)} & 100 & 91.1 & 91.9 \\
				& 500 & 93.2  & 96.2 \\
				& 1.6K & 94.8  & 98.3 \\
				& 5.2K & \textbf{95.3} & \textbf{99.6} \\				
			\end{tabular}
			\mbox{}\vspace{-.2cm}
			\subcaption{HollywoodHeads}
			\mbox{}\vspace{-.1cm}
			\begin{tabular}{c||r|cc}
				\parbox{2.6cm}{\centering Method} & Avg.\# TP & Tube Rec. & Box Rec.\\
				\hline\hline
				SS-Track (ours) & 17K & 96.8  & 98.1  \\
				Det-Track (ours) & 17K & \textbf{97.5} & 98.5  \\
				\hline
				\multirow{4}{*}{TPN (ours)} & 100 & 81.3 & 83.2  \\
				& 1K & 93.3 & 96.5 \\
				& 5K & 95.3  & 98.8 \\
				& 10K & 95.5 & \textbf{99.2} \\
				
			\end{tabular}
			\mbox{}\vspace{-.2cm}
			\subcaption{ImageNet VID}
			\mbox{}\vspace{-.8cm}
		\end{center}
		\caption{
			Evaluation of tube proposals in terms of tube-recall and box-recall on HollywoodHeads and VID datasets.
			Avg.\#TP stands for the average number of tube proposals per chunk of consecutive frames.
		}
		\label{tbl:recall_ana}
	\end{table}
	We evaluate the recall on HollywoodHeads and ImageNet VID datasets in Table~\ref{tbl:recall_ana}.
	For each dataset we select a validation set of about $3000$ frame chunks.
	We first observe that action proposals do not achieve satisfactory level of recall for object detection.
	Due the low performance, we omit action proposals from further evaluation.
	We next evaluate the two track-based tube proposals introduced in Section~\ref{sec:tube_from_track_box}.
	The proposals based on Selective Search and single-frame object detections are denoted by SS-Track and Det-Track respectively.
	Both methods obtain high recall at the cost of the large number of tube proposals.
	
	For TPN we experiment with different numbers of top-scoring tube proposals after tube-NMS.
	For HollywoodHeads TPN outperforms all other methods in both Tube recall and Box recall.
	For ImageNet VID TPN achieves slightly worse recall compared to SS-Track and Det-Track.
	This might be due to the large variation of object aspect ratio in ImageNet VID.
	Figure~\ref{fig:recall_ana} illustrates recall depending on the number of tube proposals.
	In all cases TPN achieves a significant reduction in the number of proposals by the cost of a minor drop of recall.
	This property can be used to significantly reduce computational complexity of detection at test time.

	\subsection{Detection results}
	\label{sec:det_res}
	In this section we report results for object detection.
	Following the PASCAL VOC detection challenge~\cite{everingham2010pascal}, we use Average Precision (AP) measure to evaluate object detection on the level of individual frames.
	Given results of a tube object detector, we decompose tubes into per-frame object detections.
	On each frame we aggregate detections from multiple tubes using the standard NMS procedure.
	Resulting detections are evaluated against ground truth object bounding boxes on test frames.
	
	\begin{table}[t]
		\begin{center}
			\vspace{-0.1cm}
			\begin{tabular}{c||cccc}
				Model & HH & Casa & VID\\
				\hline
				\hline
				Context-aware CNN~\cite{vu2015context} & 72.7 & 72.7 & -\\
				Kang~\cite{kang2017object} & - & - & 68.4\\
				\hline
				Box-CNN & 82.4 & 81.9 & 68.7\\
				Box-CNN + track & 84.7 & 82.1 & 68.9\\
				Tube-CNN & \textbf{86.8} & \textbf{84.0} & \textbf{72.7}
			\end{tabular}
		\end{center}
		\caption{
			Detection performance mAP ($\%$) on the HollywoodHeads, Casablanca and ImageNet VID datasets.
		}
		\label{tbl:des_performance_HH_Casa}
	\end{table}

	\begin{table*}[t]
		\begin{center}
			\begin{tabular}{@{}c||c@{\;\;\;}c@{\;\;\;}c@{\;\;\;}c@{\;\;\;}c@{\;\;\;}c@{\;\;\;}c@{\;\;\;}c@{\;\;\;}c@{\;\;\;}c@{\;\;\;}c@{\;\;\;}c@{\;\;\;}c@{\;\;\;}c@{\;\;\;}c@{\;\;\;}c@{}}
				Model&\includegraphics[width=0.4cm]{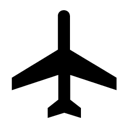}&\includegraphics[width=0.4cm]{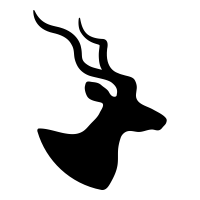}&\includegraphics[width=0.4cm]{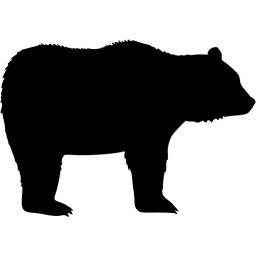}&\includegraphics[width=0.4cm]{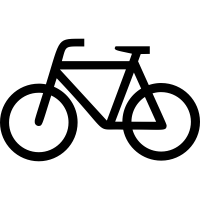}&\includegraphics[width=0.4cm]{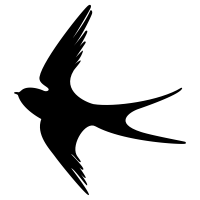}&\includegraphics[width=0.4cm]{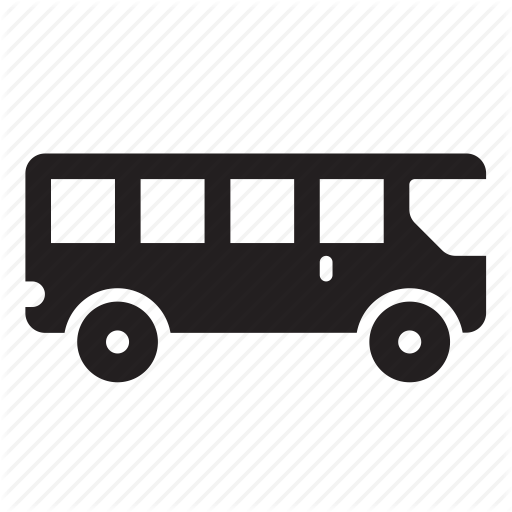}&\includegraphics[width=0.4cm]{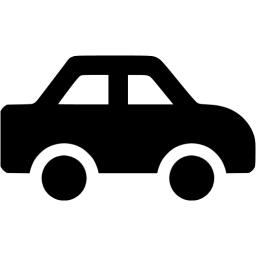}&\includegraphics[width=0.4cm]{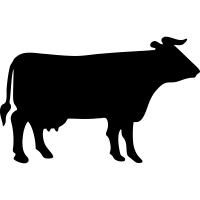}&\includegraphics[width=0.4cm]{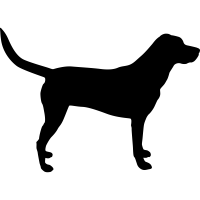}&\includegraphics[width=0.4cm]{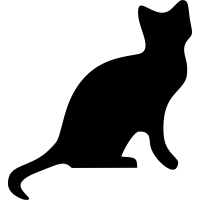}&\includegraphics[width=0.4cm]{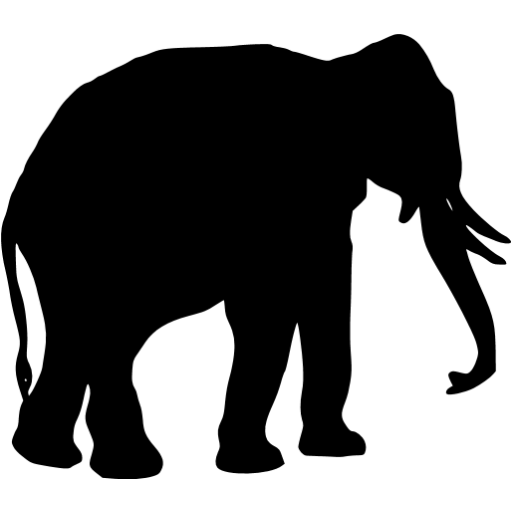}&\includegraphics[width=0.4cm]{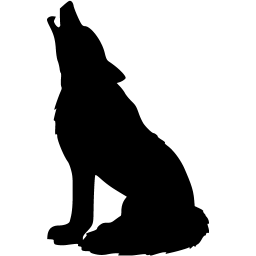}&\includegraphics[width=0.4cm]{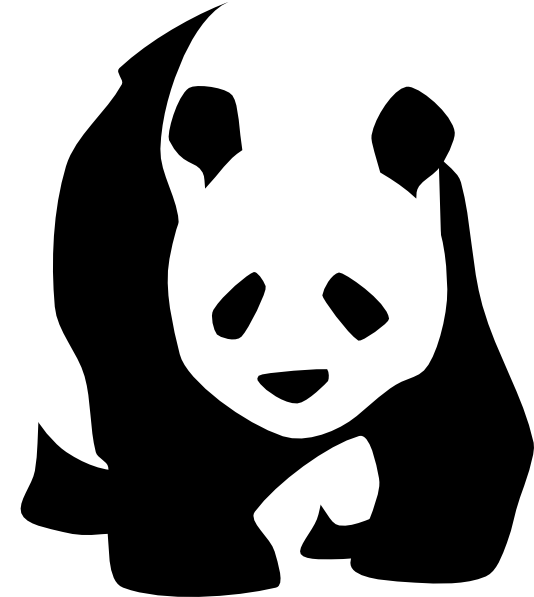}&\includegraphics[width=0.4cm]{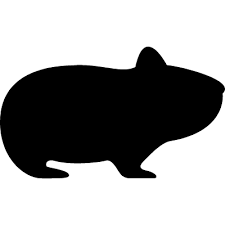}&\includegraphics[width=0.4cm]{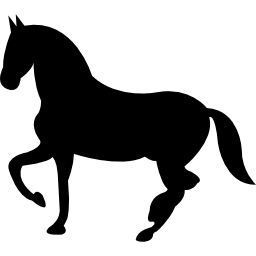}&\includegraphics[width=0.4cm]{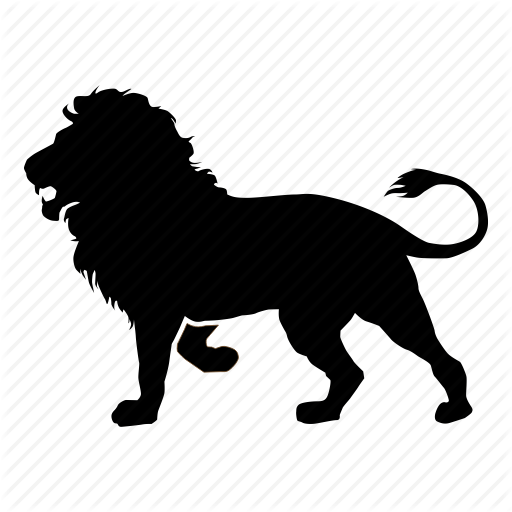}\\
				\hline\hline
				Kang~\cite{kang2017object} & \textbf{84.6} & \textbf{78.1} & 72.0 & 67.2 & 68.0 & 80.1 & 54.7 & \textbf{61.2} & 61.6 & \textbf{78.9} & 71.6 & 83.2 & 78.1 & 91.5 & 66.8 & 21.6\\
				
				Box-CNN&79.1&73.7&79&66.3&68&80.3&56.8&59.1&70.8&75.6&71.7&82.5&76.9&84.1&62.9&25.1\\
				\hline
				Tube-CNN&81.1&77&\textbf{79.9}&\textbf{72.5}&\textbf{72.5}&\textbf{84.2}&\textbf{56.2}&60.3&\textbf{77.8}&77.4&\textbf{73.5}&\textbf{83.9}&\textbf{82}&\textbf{94.9}&\textbf{71.6}&\textbf{40.5}\\
				\vspace{-0.3cm}\\
				Model&\includegraphics[width=0.4cm]{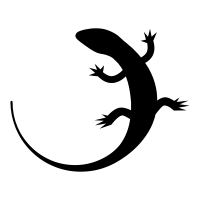}&\includegraphics[width=0.4cm]{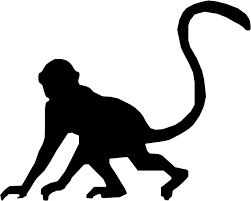}&\includegraphics[width=0.4cm]{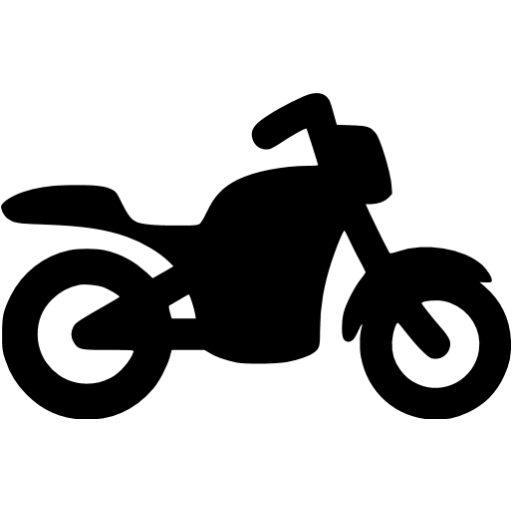}&\includegraphics[width=0.4cm]{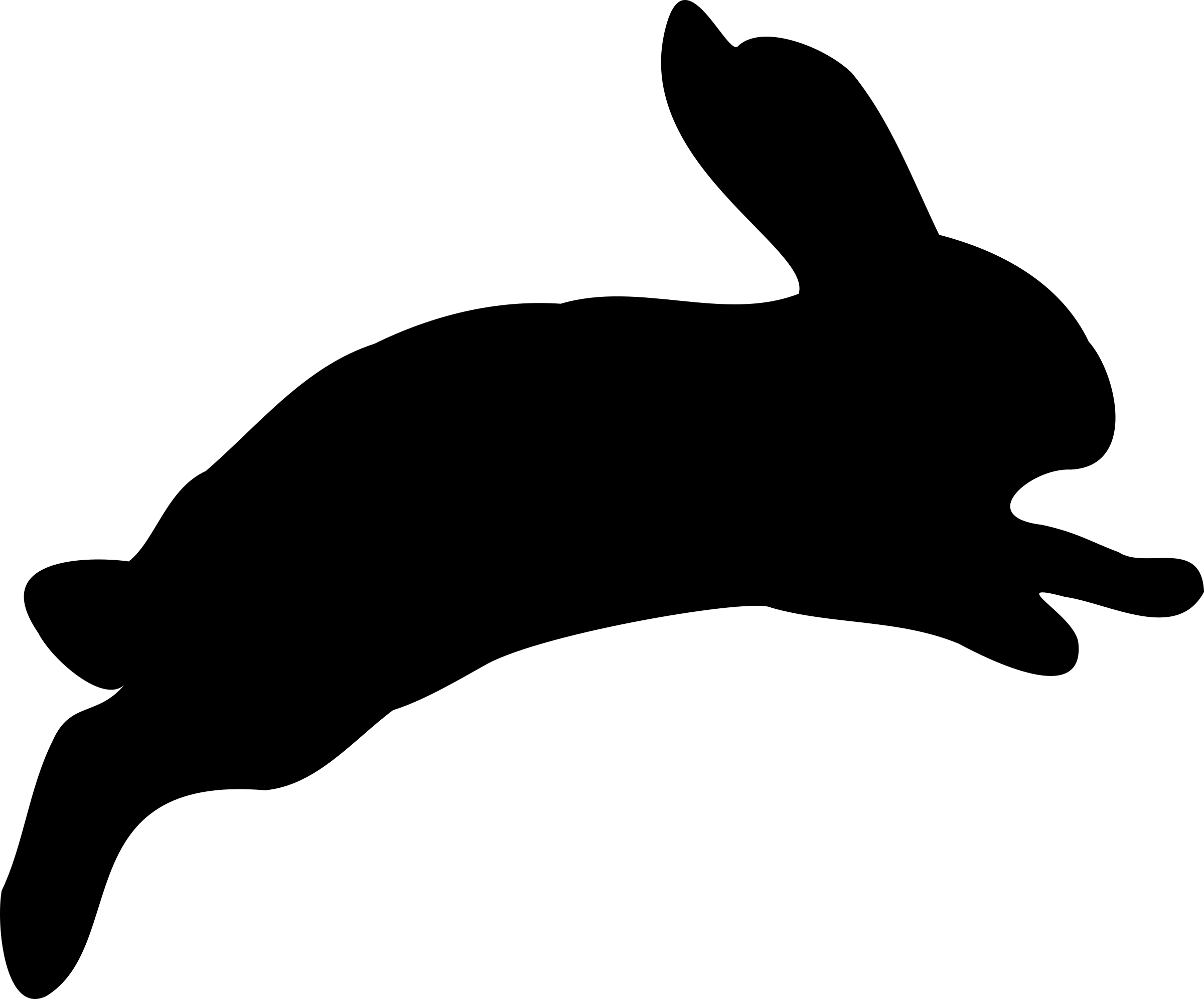}&\includegraphics[width=0.4cm]{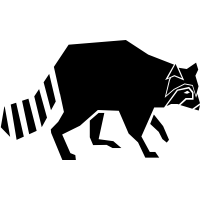}&\includegraphics[width=0.4cm]{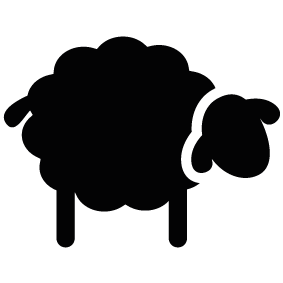}&\includegraphics[width=0.4cm]{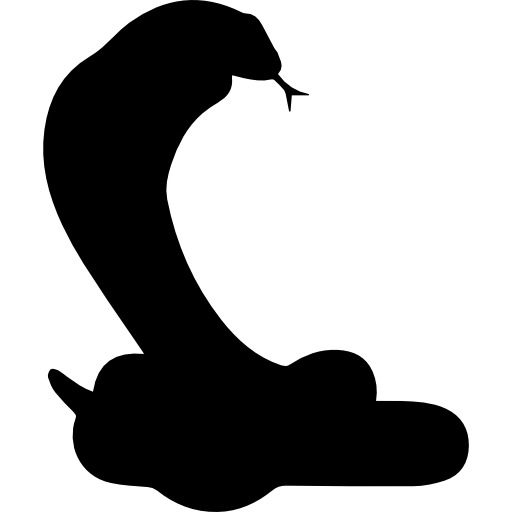}&\includegraphics[width=0.4cm]{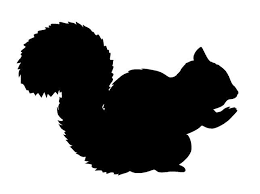}&\includegraphics[width=0.4cm]{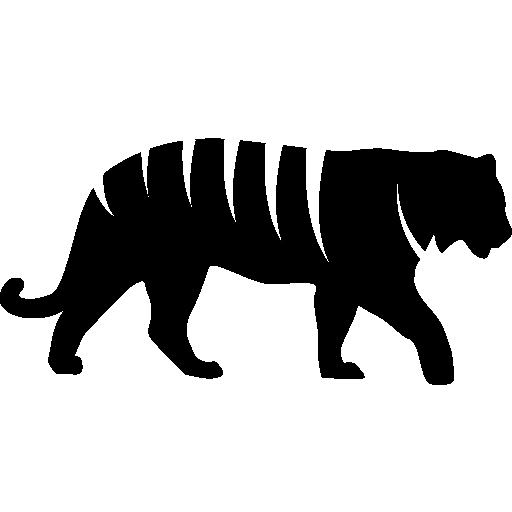}&\includegraphics[width=0.4cm]{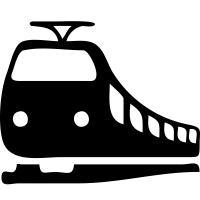}&\includegraphics[width=0.4cm]{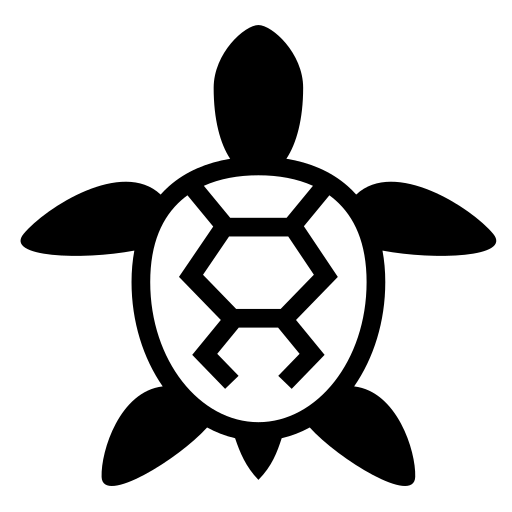}&\includegraphics[width=0.4cm]{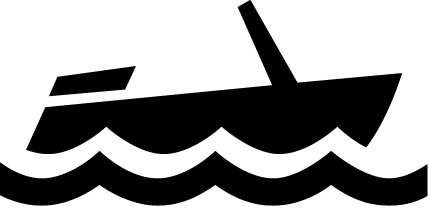}&\includegraphics[width=0.4cm]{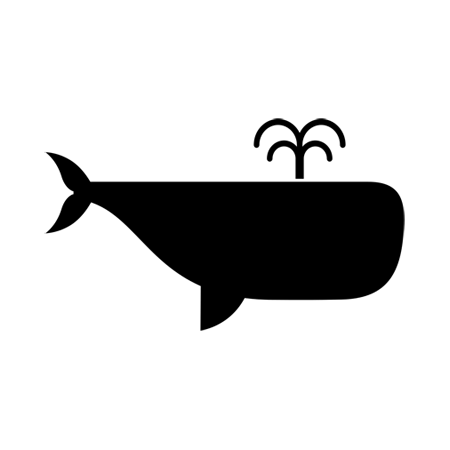}&\includegraphics[width=0.4cm]{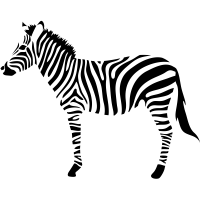}&mAP&\\
				\hline\hline
				Kang~\cite{kang2017object} & 74.4 & 36.6 & 76.3 & 51.4 & \textbf{70.6} & \textbf{64.2} & 61.2 & 42.3 & 84.8 & 78.1 & 77.2 & 61.5 & \textbf{66.9} & \textbf{88.5} & 68.4 & \\
				Box-CNN&78.3&44&76.9&58.1&53.6&56.8&72.6&52.2&88.5&78.9&77.6&64&60&87.3&68.7&\\
				\hline
				Tube-CNN & \textbf{78.8} & \textbf{50} & \textbf{83.7} & \textbf{55.5} & 56.7 & 58.5 & \textbf{75.6} & \textbf{62.0} & \textbf{90.4} & \textbf{81.4} & \textbf{80.3} & \textbf{70.7} & 65.6 & 87.4 & \textbf{72.7}&\\
			\end{tabular}
		\end{center}
		\caption{
			Object detection performance AP ($\%$) on ImageNet VID validation set.}
		\label{tbl:vid_challenge_exp}
	\end{table*}
	
	\paragraph{Baselines.}
	We compare our method to the single-frame Fast R-CNN object detectors~\cite{girshick2015fast} which we re-train on our data.
	We refer to such detectors as Box-CNN to contrast with the tube-level models.
	Box-CNNs are trained on single frames of training videos and frame-level annotations.
	
	Single-frame object detectors often produce noisy detections along object tracks~\cite{kang2016object}.
	A common practice to improve object detection in video is to link detections on consecutive frames with tracking-by-detection methods~\cite{everingham2006hello,pirsiavash2011globally}.
	We here follow the procedure in~\cite{everingham2006hello} and link object detections across frames based on the consensus of KLT point tracks.
	This procedure enables to overcome some failures of object detectors by bridging the gaps in object tracks along time and by discarding short tracks.
	We apply tracking-by-detection to the Box-CNN method and denote this strong baseline as "Box-CNN + track".
	
\begin{figure*}[]
		\begin{center}
			\begin{tabular}{@{}c@{\,}c@{\,}c@{\,}c@{}}
				\multicolumn{2}{c}{\textit{Box-CNN}}&\multicolumn{2}{c}{\textit{Tube-CNN}}\\[0.1cm]
				\multicolumn{2}{c}{\hspace{-0.19cm}\includegraphics[trim = 0mm 30mm 20mm 15mm, clip, width=0.5\linewidth]{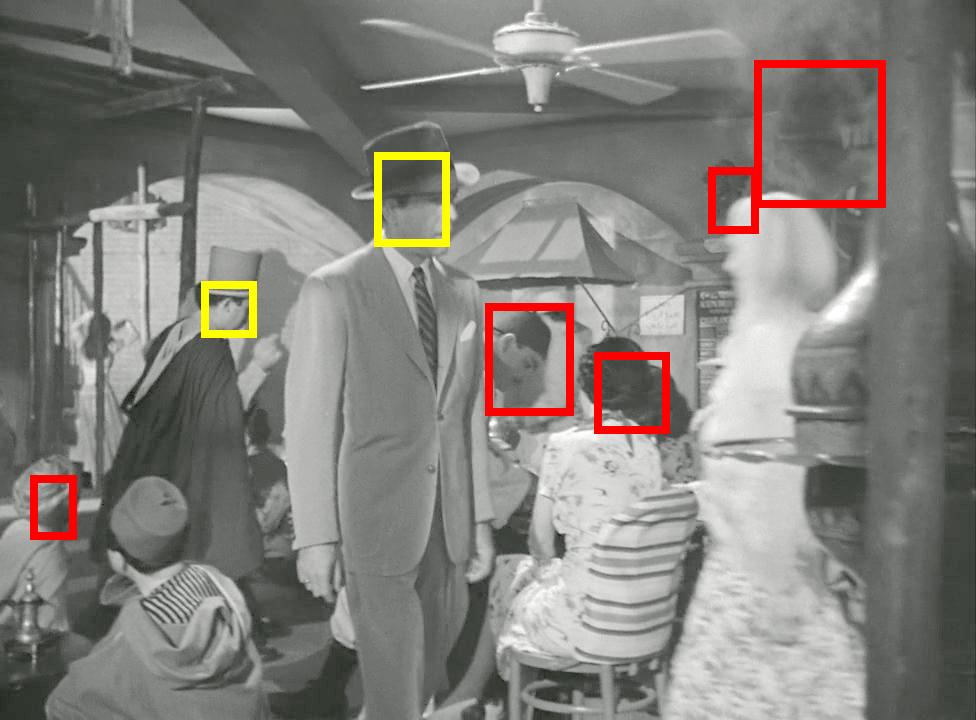}}
				&
				\multicolumn{2}{c}{\hspace{-0.19cm}\includegraphics[trim = 0mm 30mm 20mm 15mm, clip, width=0.5\linewidth]{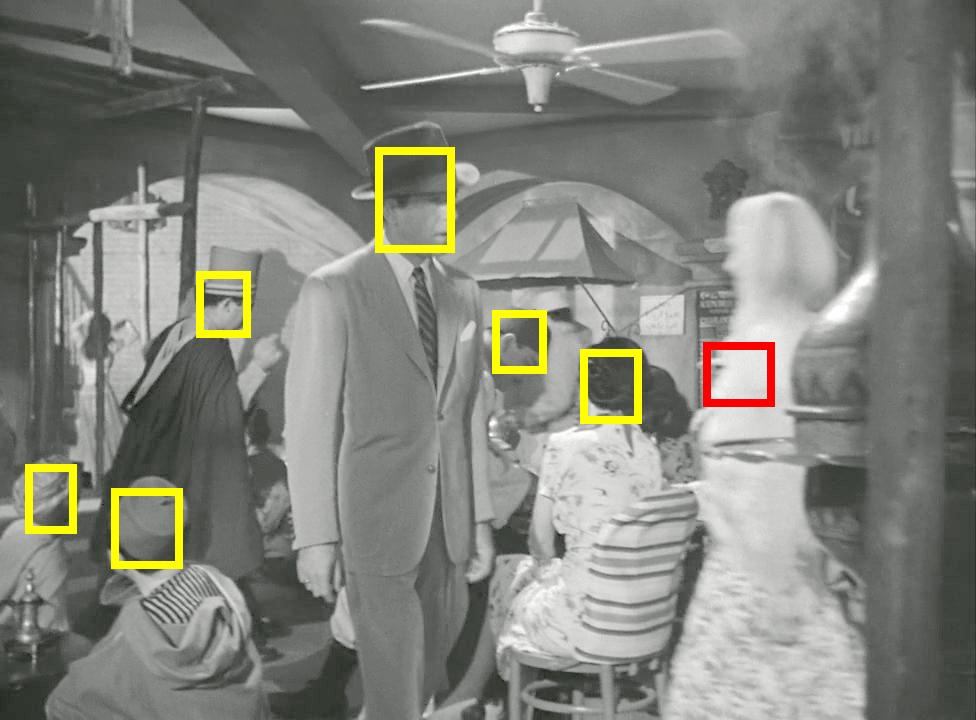}}\vspace{-0.08cm}\\
				\textit{Box-CNN} & \textit{Tube-CNN} & \textit{Box-CNN} & \textit{Tube-CNN} \\
				\includegraphics[trim = 0mm 40mm 30mm 15mm, clip, width=0.25\linewidth]{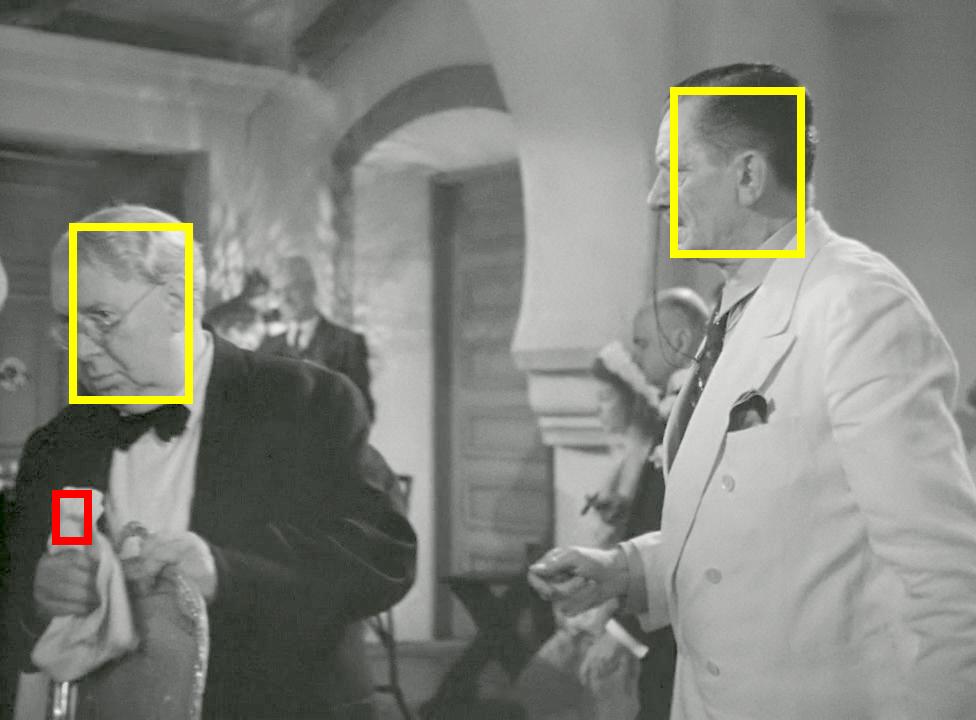}
				&
				\hspace{-.17cm}\includegraphics[trim = 0mm 40mm 30mm 15mm, clip, width=0.25\linewidth]{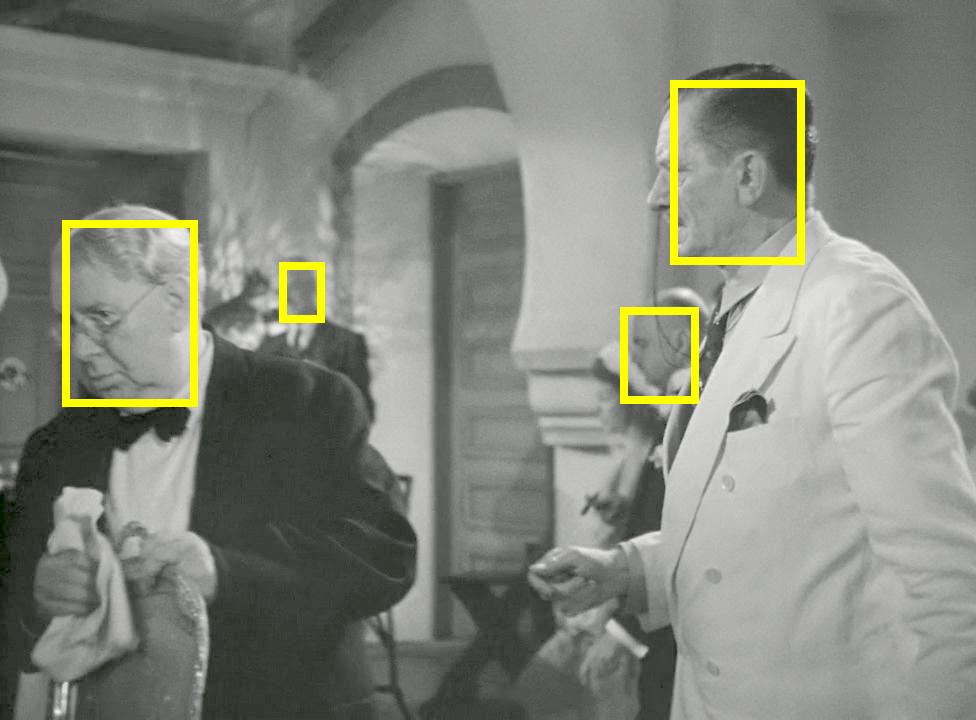}
				&
				\hspace{-0cm}\includegraphics[trim = 80mm 80mm 100mm 70mm, clip, width=0.25\linewidth]{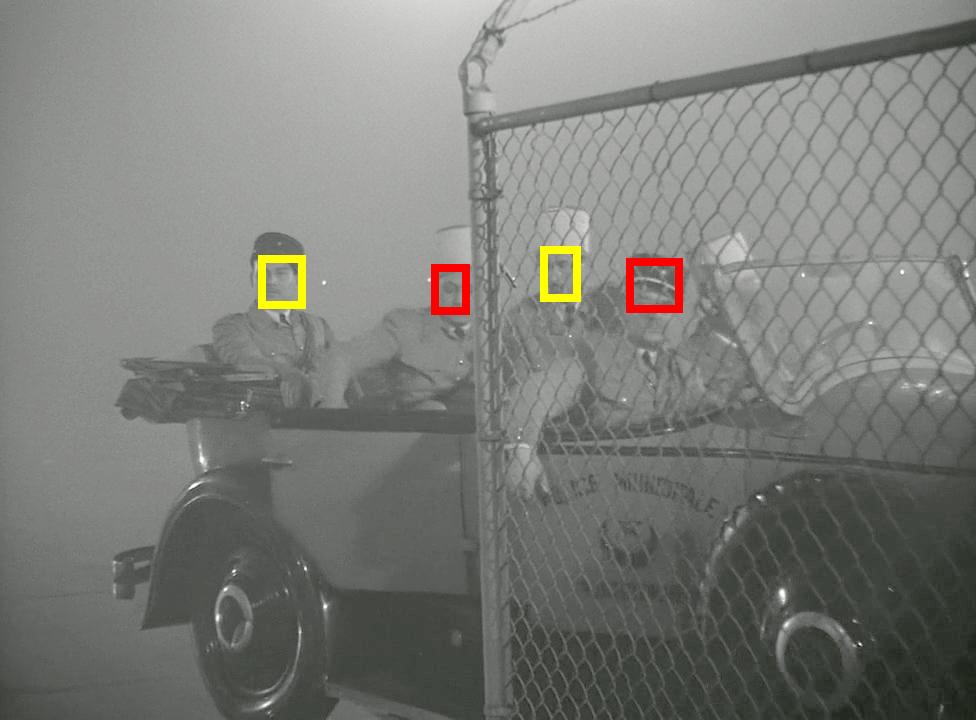}
				&
				\hspace{-0.24cm}\includegraphics[trim = 80mm 80mm 100mm 70mm, clip, width=0.25\linewidth]{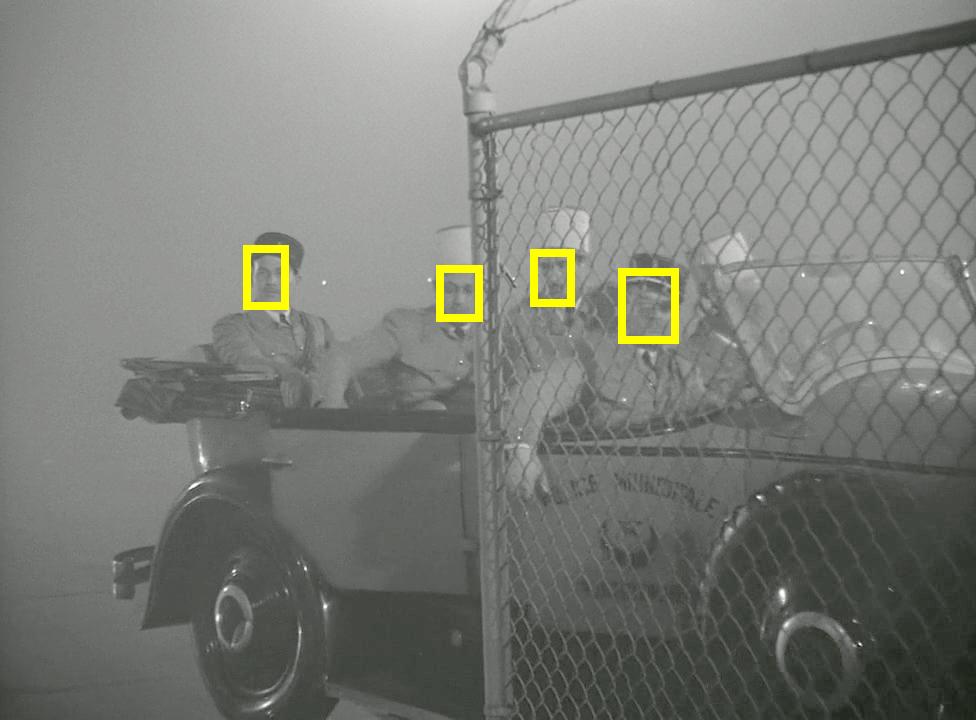}\vspace{-0.08cm}\\
				\includegraphics[trim = 0mm 40mm 0mm 36mm, clip, width=0.25\linewidth]{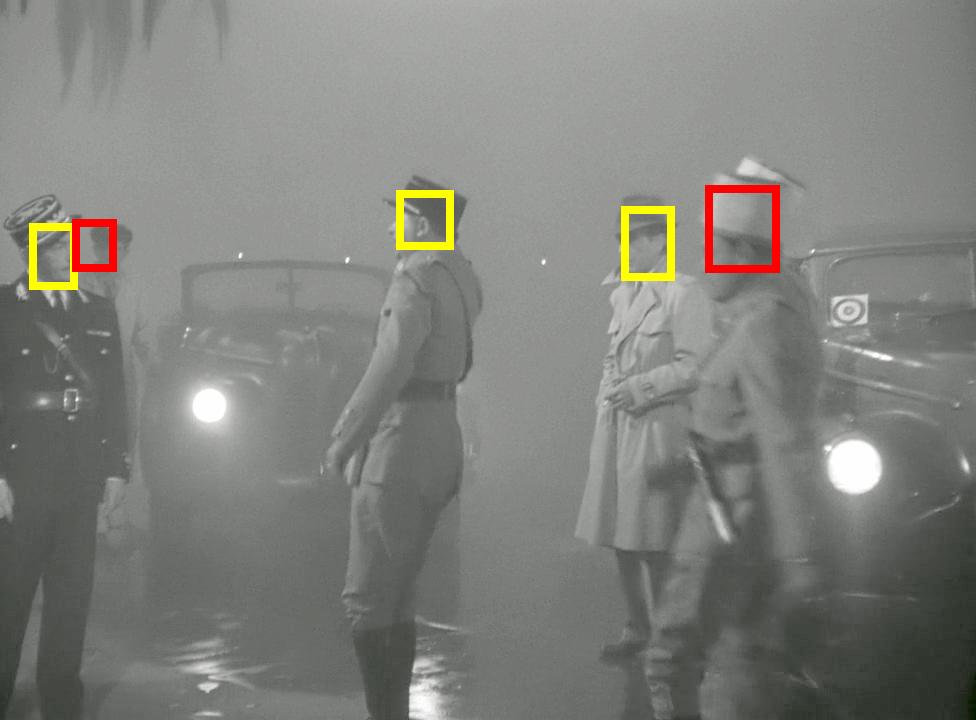}
				&
				\hspace{-0.17cm}\includegraphics[trim = 0mm 40mm 0mm 36mm, clip, width=0.25\linewidth]{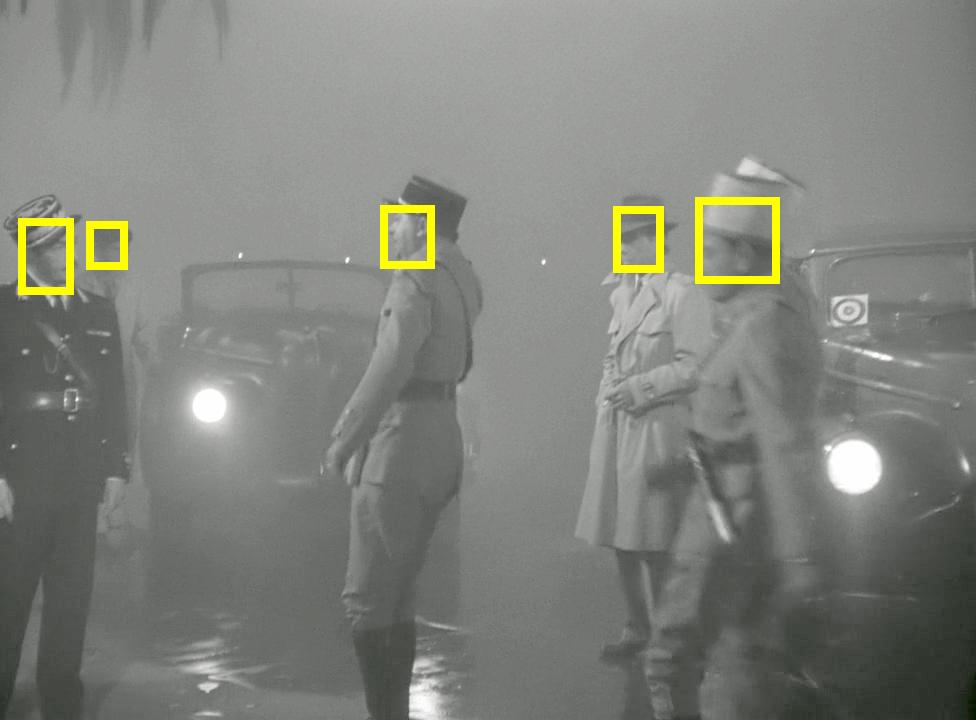}
				&
				\hspace{-0cm}\includegraphics[trim = 0mm 40mm 0mm 36mm, clip, width=0.25\linewidth]{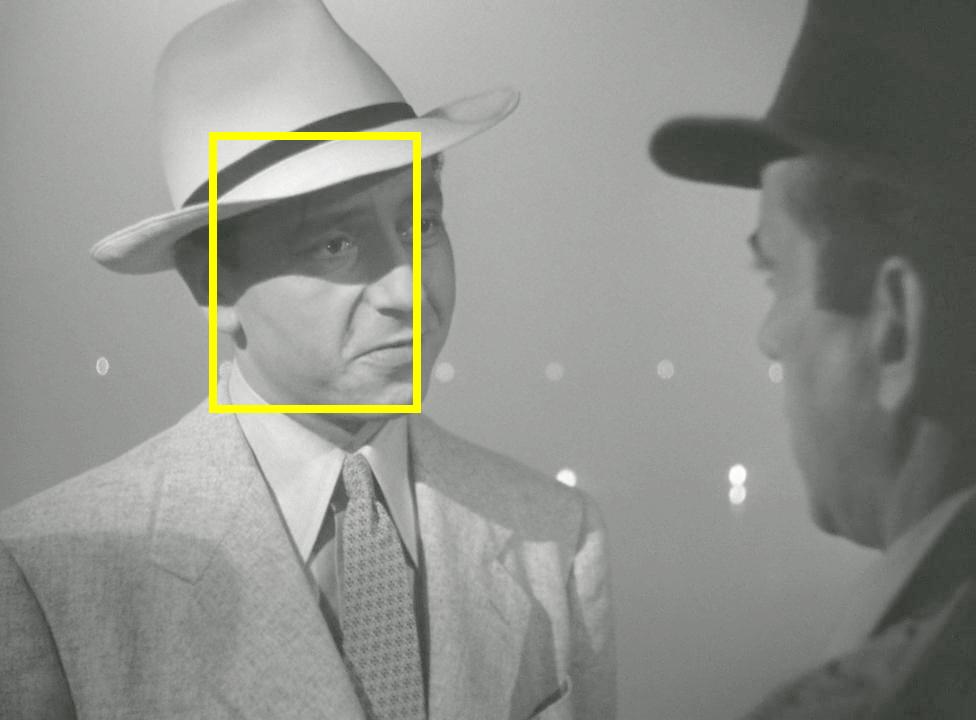}
				&
				\hspace{-0.24cm}\includegraphics[trim = 0mm 40mm 0mm 36mm, clip, width=0.25\linewidth]{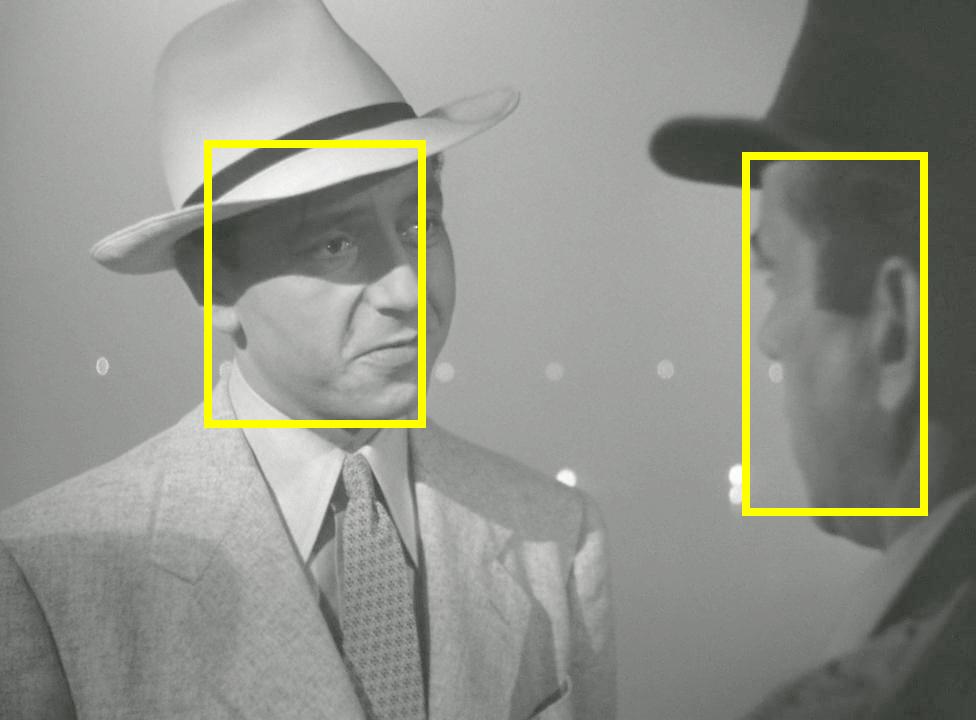}\\
				
				\includegraphics[trim = 0mm 95mm 65mm 0mm, clip, width=0.25\linewidth]{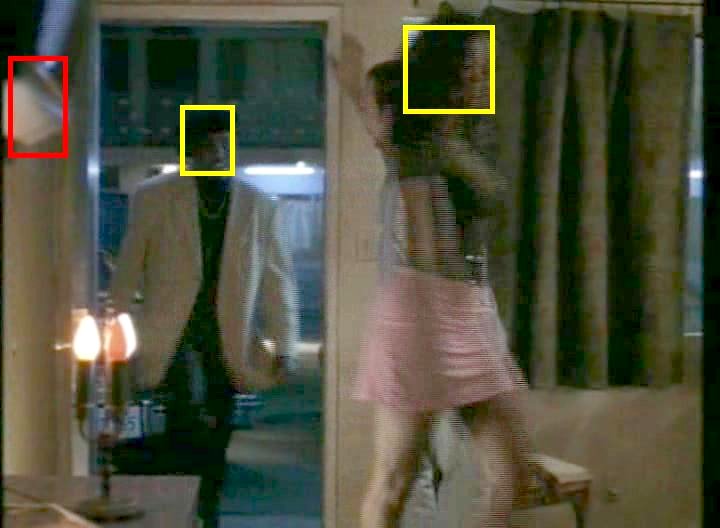}
				&
				\hspace{-0.17cm}\includegraphics[trim = 0mm 95mm 65mm 0mm, clip, width=0.25\linewidth]{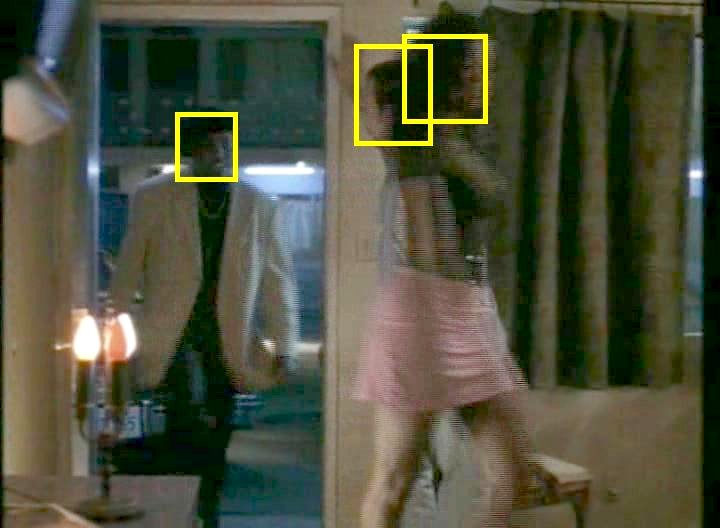}
				&
				\hspace{0.cm}\includegraphics[trim = 40mm 38mm 20mm 0mm, clip, width=0.25\linewidth]{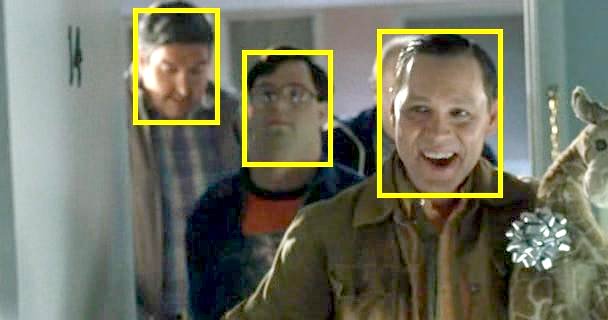}
				&
				\hspace{-0.24cm}\includegraphics[trim = 40mm 38mm 20mm 0mm, clip, width=0.25\linewidth]{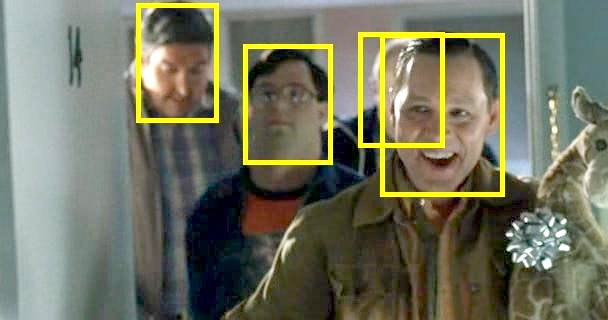}\vspace{-0.08cm}\\
				\includegraphics[trim = 40mm 40mm 50mm 30mm, clip, width=0.25\linewidth]{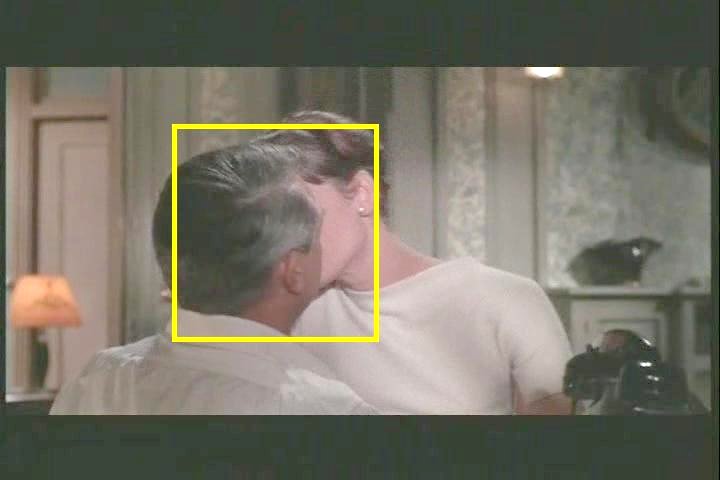}
				&
				\hspace{-0.17cm}\includegraphics[trim = 40mm 40mm 50mm 30mm, clip, width=0.25\linewidth]{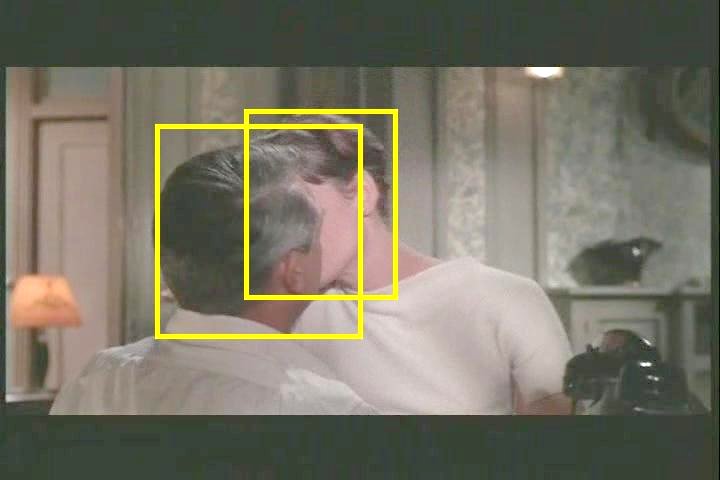}
				&
				\hspace{0cm}\includegraphics[trim = 27mm 0mm 0mm 0mm, clip, width=0.25\linewidth]{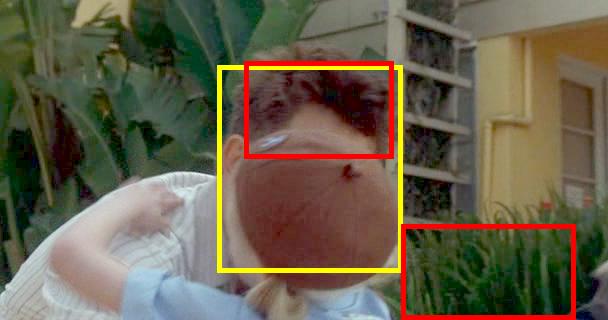}
				&
				\hspace{-0.24cm}\includegraphics[trim = 27mm 0mm 0mm 0mm, clip, width=0.25\linewidth]{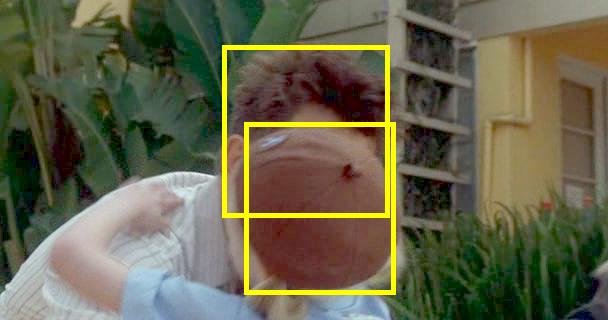}\vspace{-0.08cm}\\
				\includegraphics[trim = 0mm 0mm 0mm 0mm, clip, width=0.25\linewidth]{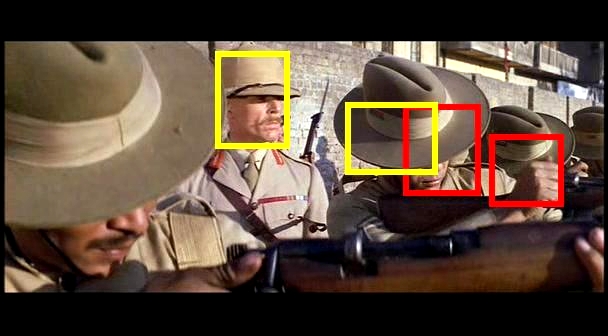}
				&
				\hspace{-0.17cm}\includegraphics[trim = 0mm 0mm 0mm 00mm, clip, width=0.25\linewidth]{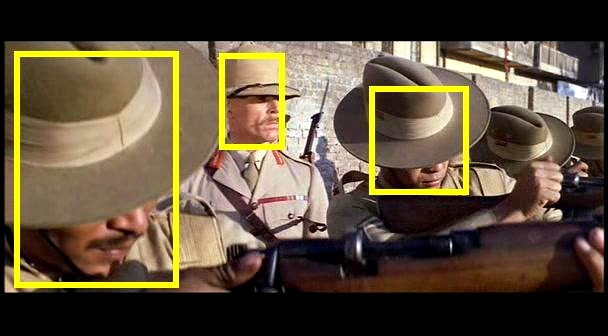}
				&
				\hspace{0cm}\includegraphics[trim = 100mm 40.2mm 0mm 15mm, clip, width=0.25\linewidth]{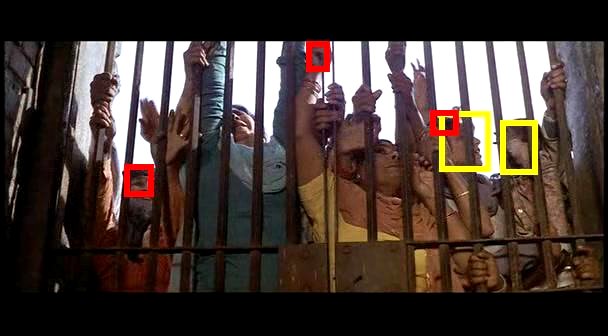}
				&
				\hspace{-0.24cm}\includegraphics[trim = 100mm 40.2mm 0mm 15mm, clip, width=0.25\linewidth]{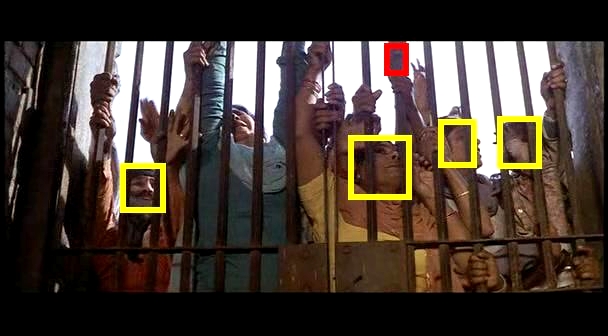}\\
				
				\includegraphics[trim = 0mm 0mm 0mm 0mm, clip, width=0.25\linewidth]{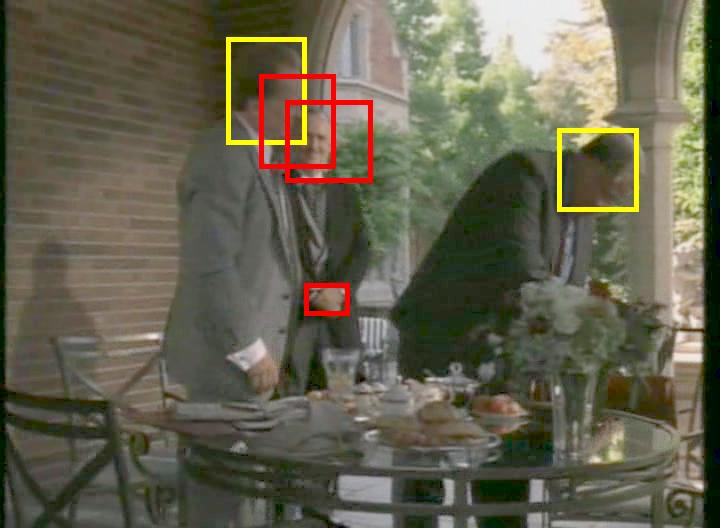}
				&
				\hspace{-0.17cm}\includegraphics[trim = 0mm 0mm 0mm 00mm, clip, width=0.25\linewidth]{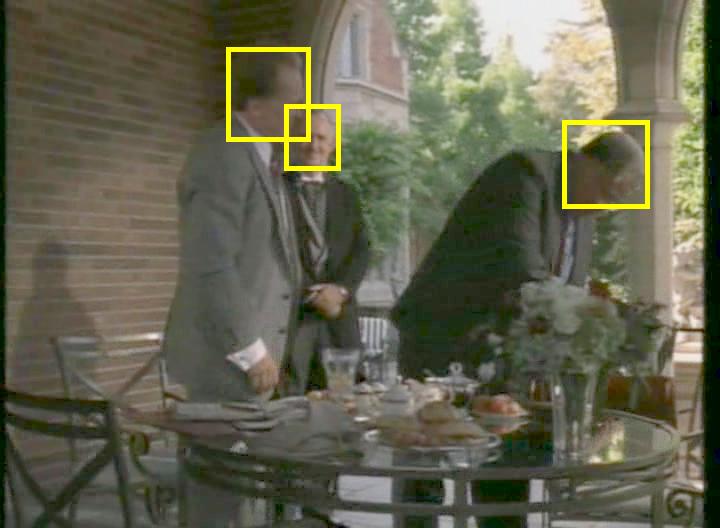}
				&
				\hspace{0cm}\includegraphics[trim = 0mm 0mm 0mm 0mm, clip, width=0.25\linewidth]{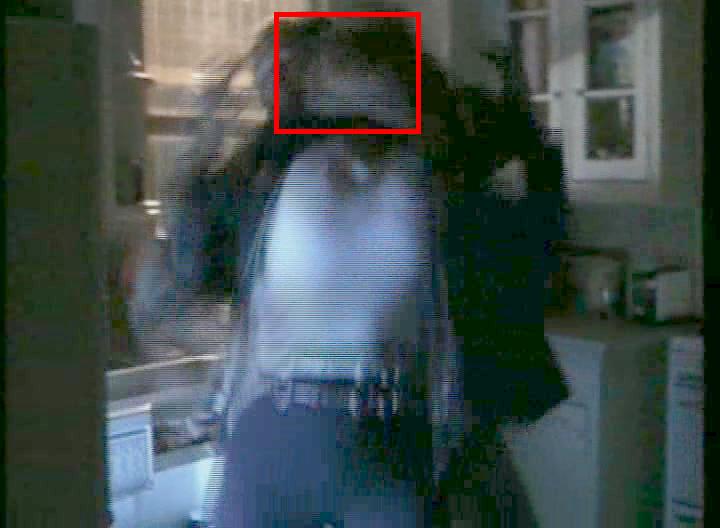}
				&
				\hspace{-0.24cm}\includegraphics[trim = 0mm 0mm 0mm 0mm, clip, width=0.25\linewidth]{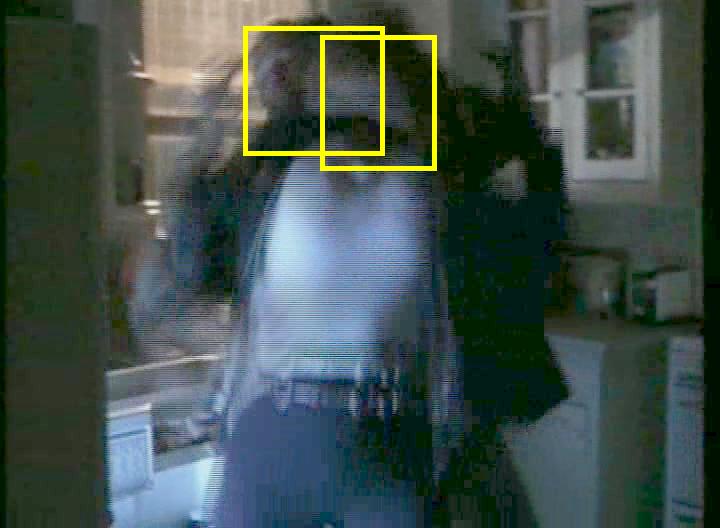}
			\end{tabular}
		\end{center}
		\caption{Qualitative results for the Tube-CNN and Box-CNN models on Casablanca dataset and the HollywoodHeads-Hard set.
			On each frame we illustrate results of the Box-CNN model and the Tube-CNN model.
			For both methods we choose thresholds such that precision equals recall on the corresponding dataset.
			All detections with scores above threshold are plotted.
			Yellow boxes: correct detections, red boxes: false detections.}
		\label{fig:supmat_qual_res_casa}
	\end{figure*}

	\paragraph{Head detection result.}
	In the first and second columns of Table~\ref{tbl:des_performance_HH_Casa} we compare our method to the state-of-the-art~\cite{vu2015context} as well as to Box-CNN and Box-CNN + track baselines on HollywoodHeads and Casablanca datasets.
	Our Tube-CNN significantly outperforms~\cite{vu2015context} by more than 10\% AP on both datasets.
	As will be detailed in Section~\ref{sec:discuss} this improvement originates both from the Tube-CNN object detectors and from the more powerful ResNet-101 base network.
	Our best Tube-CNN model is trained on SS-Track tube proposals.
	Tube-CNN achieves consistent improvement over other baseline methods using comparable base CNN architectures.
	\begin{figure}[t]
		\includegraphics[trim = 0cm 0cm 0cm 0cm, clip, width=0.42\textwidth]{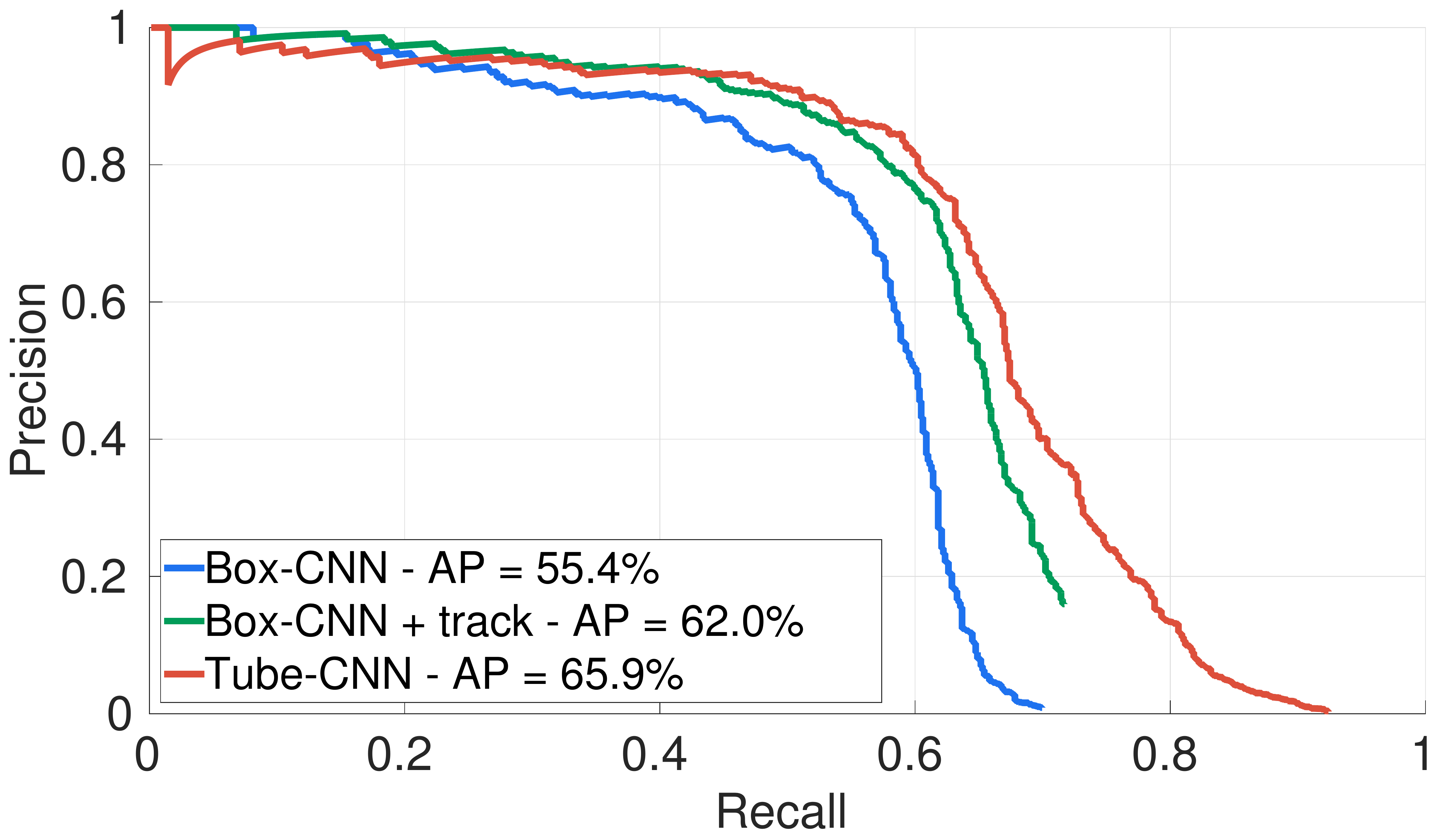}
		\caption{Detection results on HollywoodHeads-Hard.}
		\label{fig:ap_hard_case}
	\end{figure}
	Figure~\ref{fig:ap_hard_case} illustrates results for the HollywoodHeads-Hard test set with occluded heads.
	Compared to the full test set, the performance on HollywoodHeads-Hard is significantly lower for all tested methods.
	Compared to the Box-CNN, Tube-CNN shows improvement of almost 10\% AP, confirming the advantage of our method for particularly difficult scenes.
	Qualitative results of Tube-CNN for difficult examples of occluded heads are illustrated in Figure~\ref{fig:supmat_qual_res_casa}.

	\begin{table}[t]
		\begin{center}    
			\scriptsize
			\hspace*{-0.4cm}
			\begin{tabular}{@{}c@{\;\;\;}||c@{\;\;\;}c@{\;\;\;}c@{\;\;\;}c@{\;\;\;}c@{\;\;\;}c@{\;\;\;}c@{\;\;\;}c@{\;\;\;}c@{\;\;\;}c@{\;\;\;}|c@{}}
				Method & \includegraphics[width=0.3cm]{figures/icons/airplane.png} & \includegraphics[width=0.3cm]{figures/icons/bird.png} & \includegraphics[width=0.3cm]{figures/icons/watercraft.png} & \includegraphics[width=0.3cm]{figures/icons/car.png} & \includegraphics[width=0.3cm]{figures/icons/cat.png} & \includegraphics[width=0.3cm]{figures/icons/cattle.png} & \includegraphics[width=0.3cm]{figures/icons/dog.png} & \includegraphics[width=0.3cm]{figures/icons/horse.png} & \includegraphics[width=0.3cm]{figures/icons/motorcycle.png} & \includegraphics[width=0.3cm]{figures/icons/train.png} & Avg. \\
				\hline
				\hline
				Kang~\cite{kang2016object} & 94.1 & 69.7 & 88.2 & 79.3 & 76.6 & 18.6 & 89.6 & 89.0 & 87.3 & 75.3 & 76.8 \\
				Galteri~\cite{galteri2017spatio} & 87.8 & 94.8 & 81.7 & 95.1 & 84.3 & 97.5 & 78.0 & 61.0 & 94.8 & 76.8 & 85.2 \\
				Kang~\cite{kang2017object} & 91.2 & 99.4 & \textbf{93.1} & 94.8 & 94.3 & 99.3 & 90.2 & 87.8 & 89.7 & 84.2 & 92.4 \\
				\hline
				Tube-CNN & \textbf{98.2} & \textbf{100} & 92.4 & \textbf{97.5} & \textbf{96.5} & \textbf{99.3} & \textbf{92.1} & \textbf{96.1} & \textbf{95.8} & \textbf{89.9} & \textbf{95.8}
			\end{tabular}
			\vspace{-0.5cm}
		\end{center}
		\caption{
			Localization performance CorLoc ($\%$) on the YTO dataset.}
		\label{tbl:corloc_yto}
	\end{table}
	
\begin{figure*}[!th]
		\renewcommand{\arraystretch}{0.1}
		\begin{center}
			\begin{tabular}{@{}c@{\,}c@{\,}c@{\,}c@{}}
				\textit{Box-CNN} & \textit{Tube-CNN} & \textit{Box-CNN} & \textit{Tube-CNN} \\[0.1cm]
				\includegraphics[trim = 0mm 0mm 100mm 0mm, clip, width=0.25\linewidth]{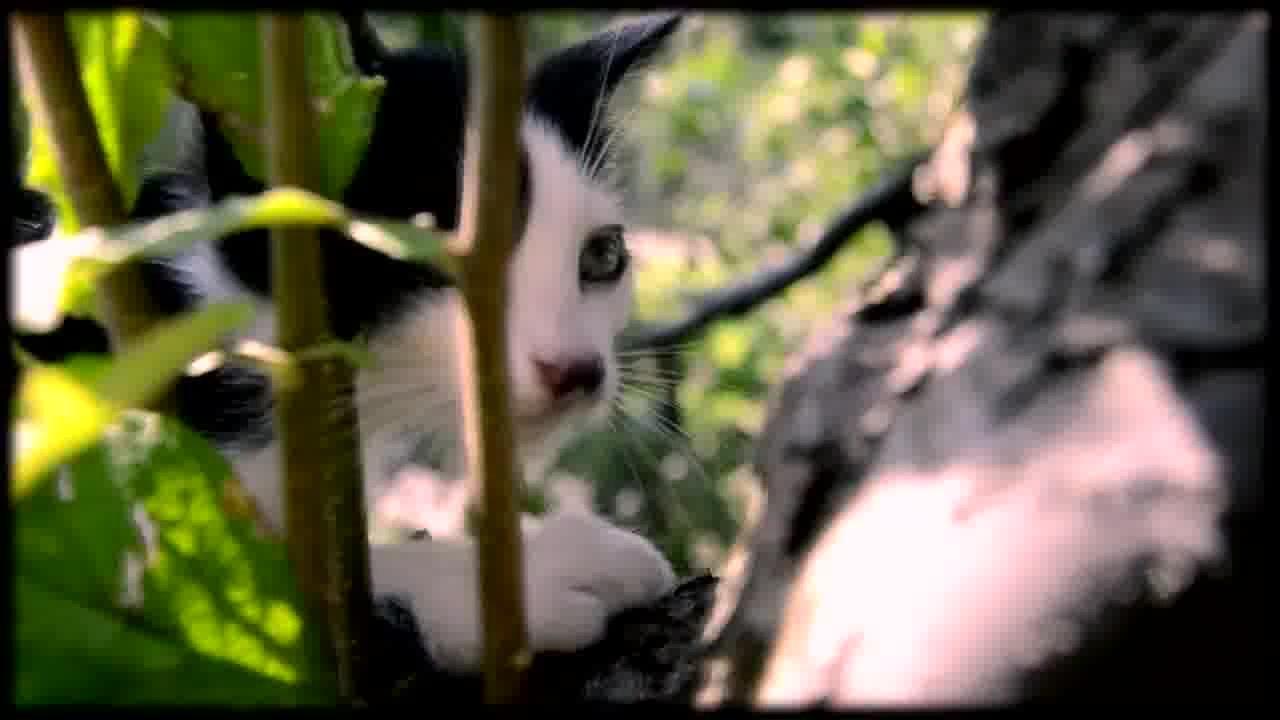}
				&
				\hspace{-0.03cm}\includegraphics[trim = 0mm 0mm 100mm 0mm, clip, width=0.25\linewidth]{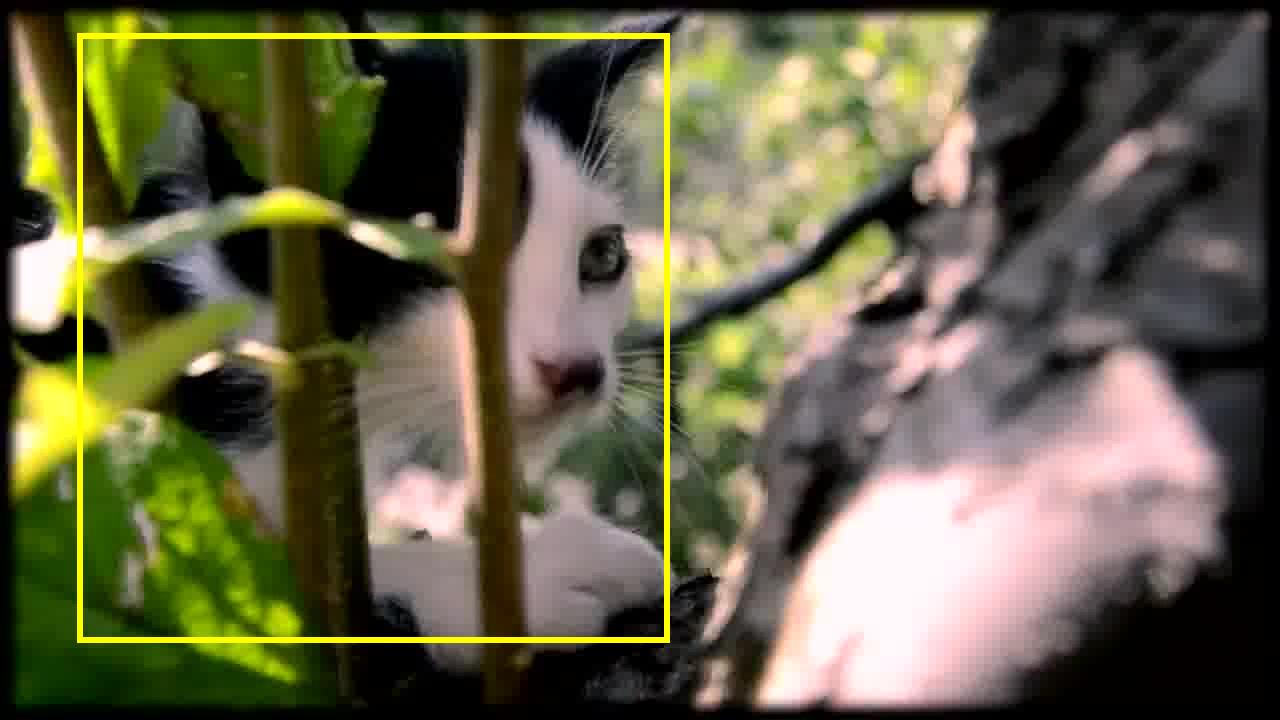}
				&
				\hspace{0.14cm}\includegraphics[trim = 120mm 20mm 0mm 15mm, clip, width=0.25\linewidth]{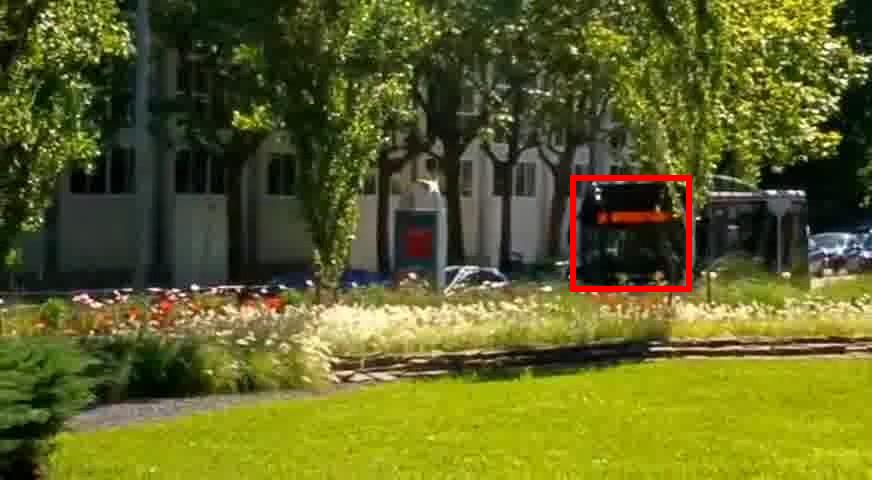}
				&
				\hspace{-0.03cm}\includegraphics[trim = 120mm 20mm 0mm 15mm, clip, width=0.25\linewidth]{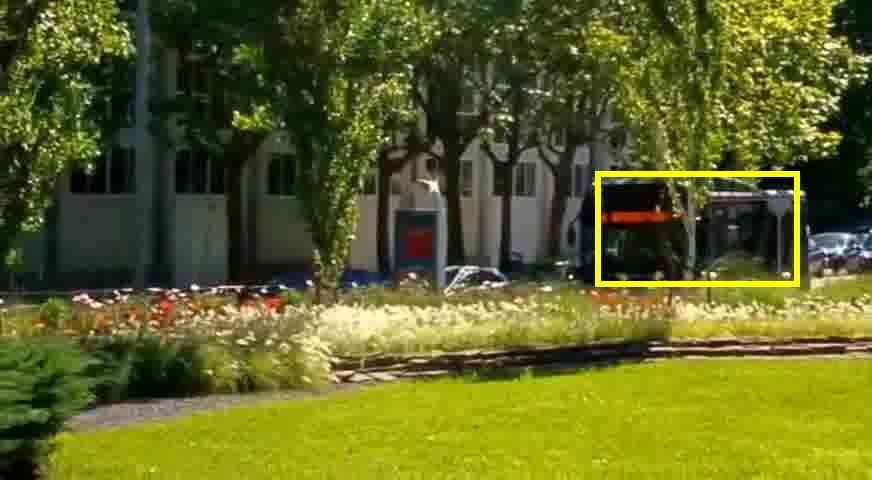}\\
				\includegraphics[trim = 0mm 40mm 150mm 40mm, clip, width=0.25\linewidth]{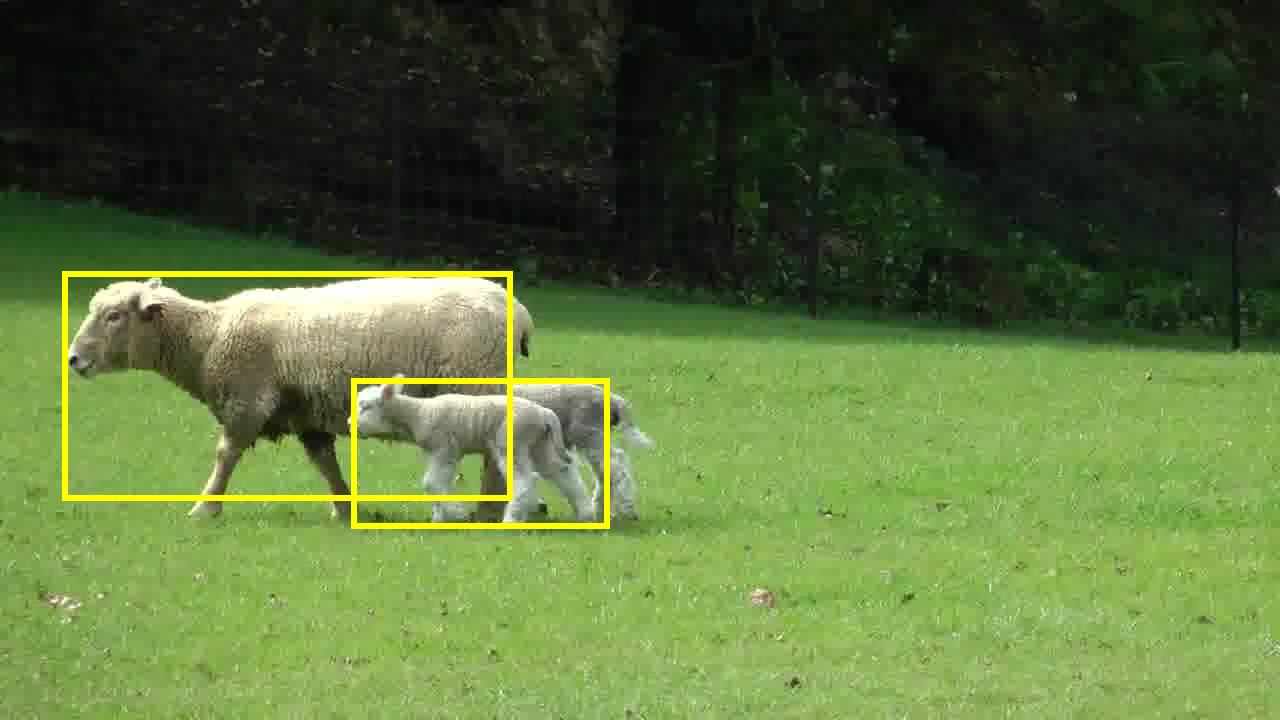}
				&
				\hspace{-0.03cm}\includegraphics[trim = 0mm 40mm 150mm 40mm, clip, width=0.25\linewidth]{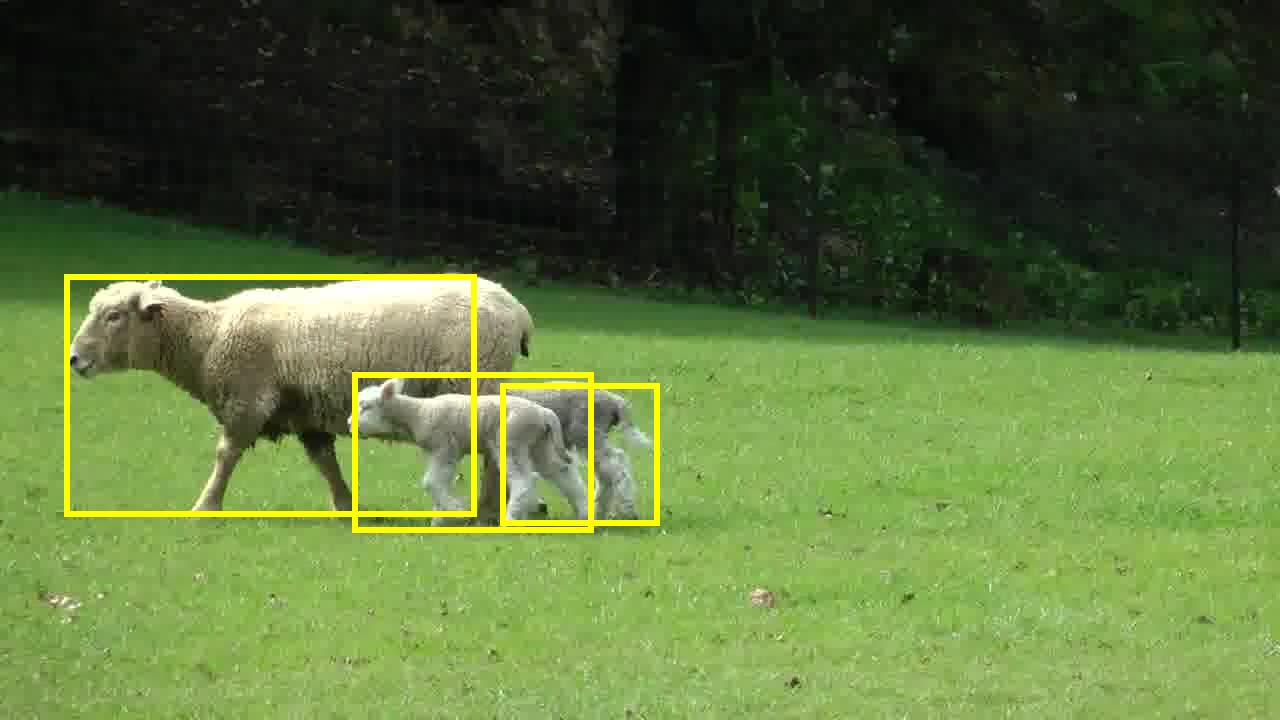}
				&
				\hspace{0.14cm}\includegraphics[trim = 30mm 0mm 80mm 55mm, clip, width=0.25\linewidth]{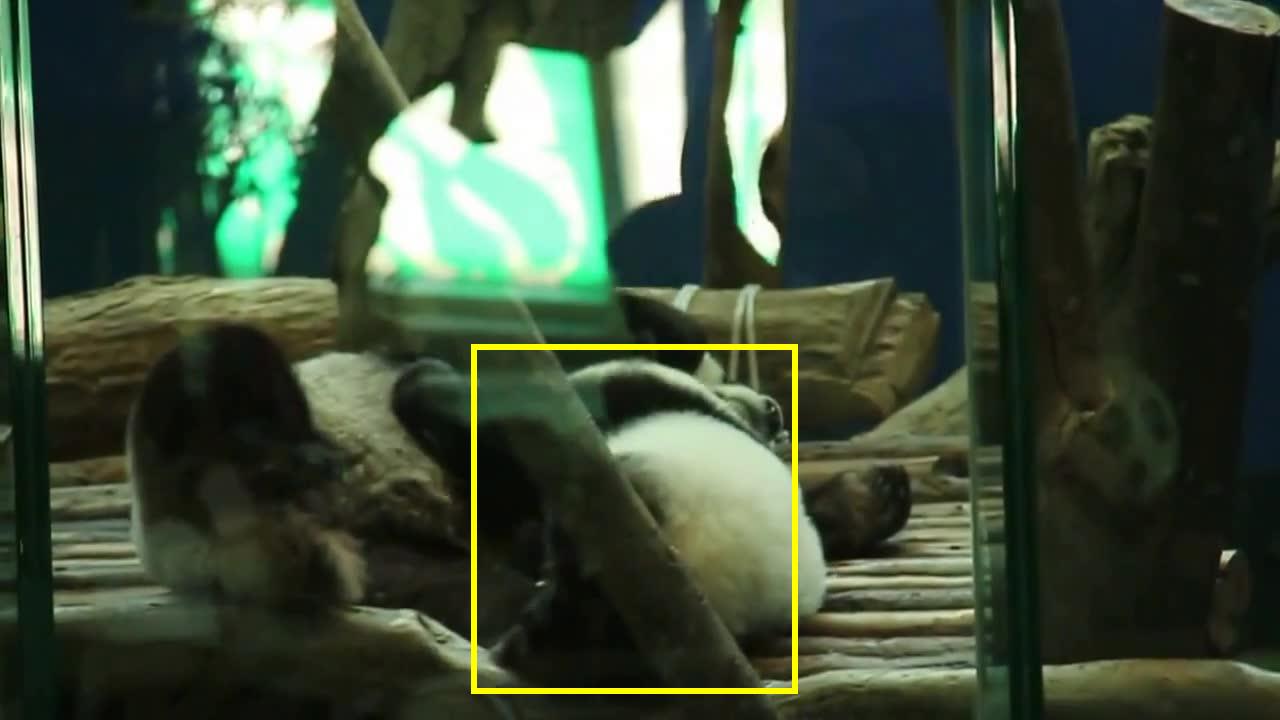}
				&
				\hspace{-0.03cm}\includegraphics[trim = 30mm 0mm 80mm 55mm, clip, width=0.25\linewidth]{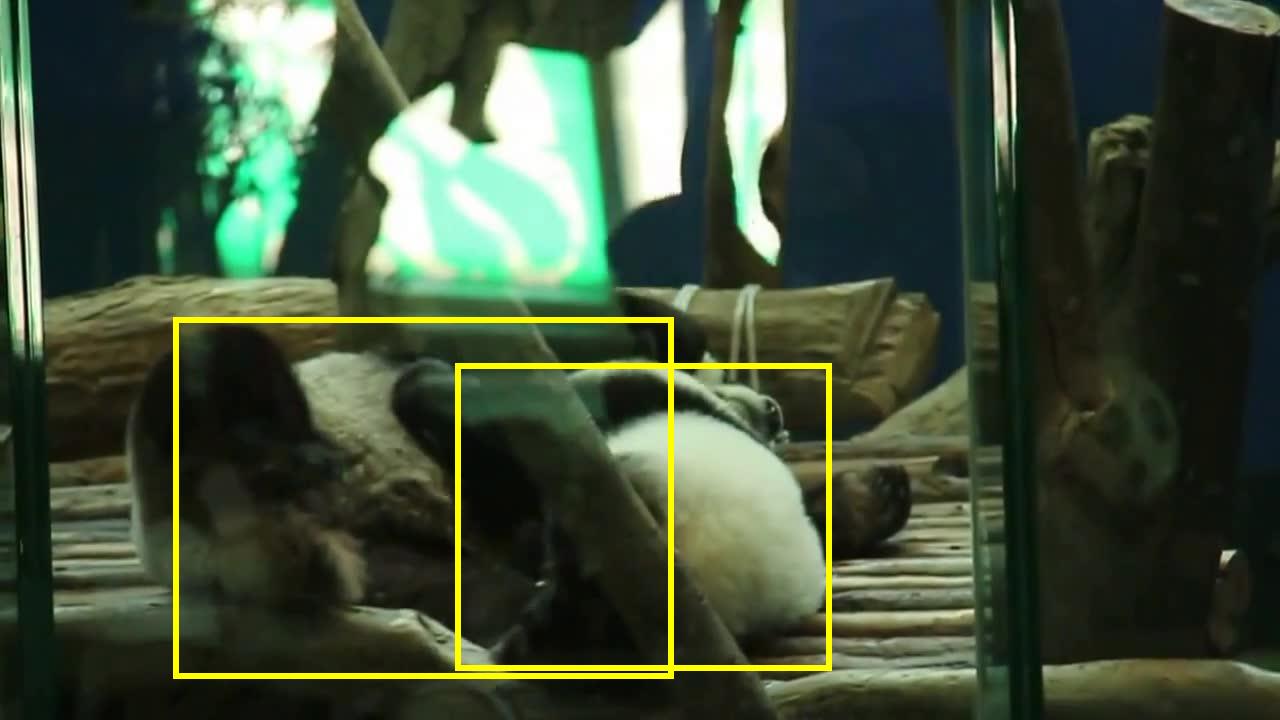}\\
				\includegraphics[trim = 0mm 0mm 0mm 0mm, clip, width=0.25\linewidth]{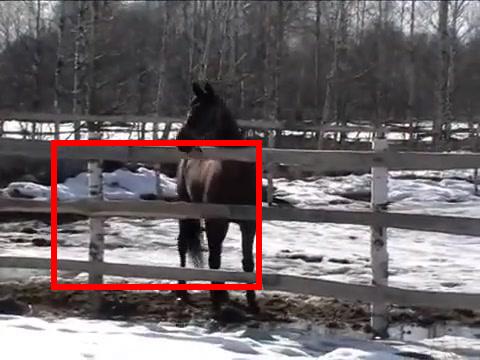}
				&
				\hspace{-0.03cm}\includegraphics[trim = 0mm 0mm 0mm 0mm, clip, width=0.25\linewidth]{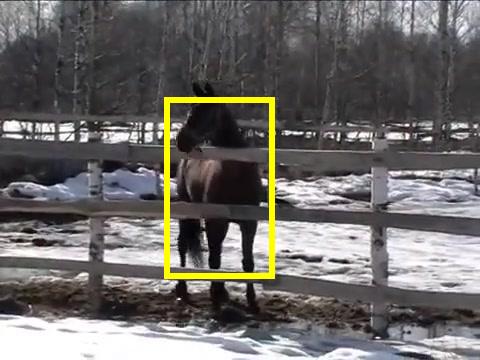}
				&
				\hspace{0.14cm}\includegraphics[trim = 115mm 0mm 0mm 0mm, clip, width=0.25\linewidth]{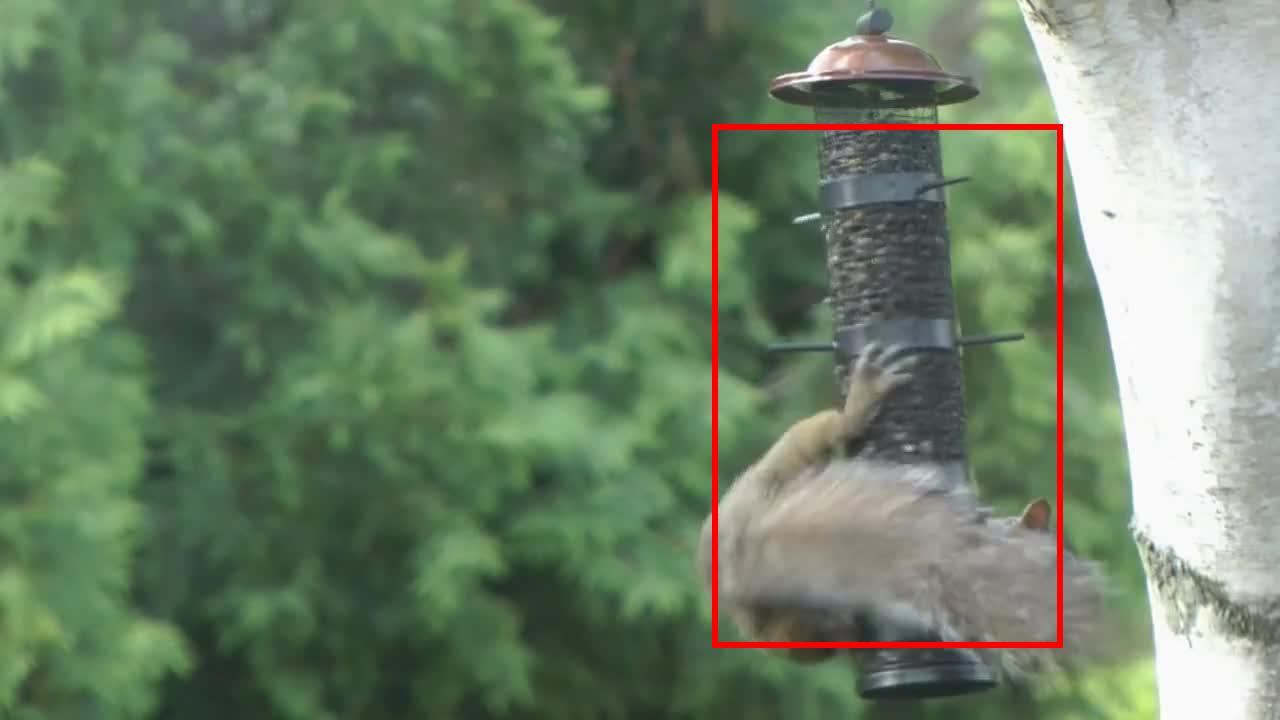}
				&
				\hspace{-0.03cm}\includegraphics[trim = 115mm 0mm 0mm 0mm, clip, width=0.25\linewidth]{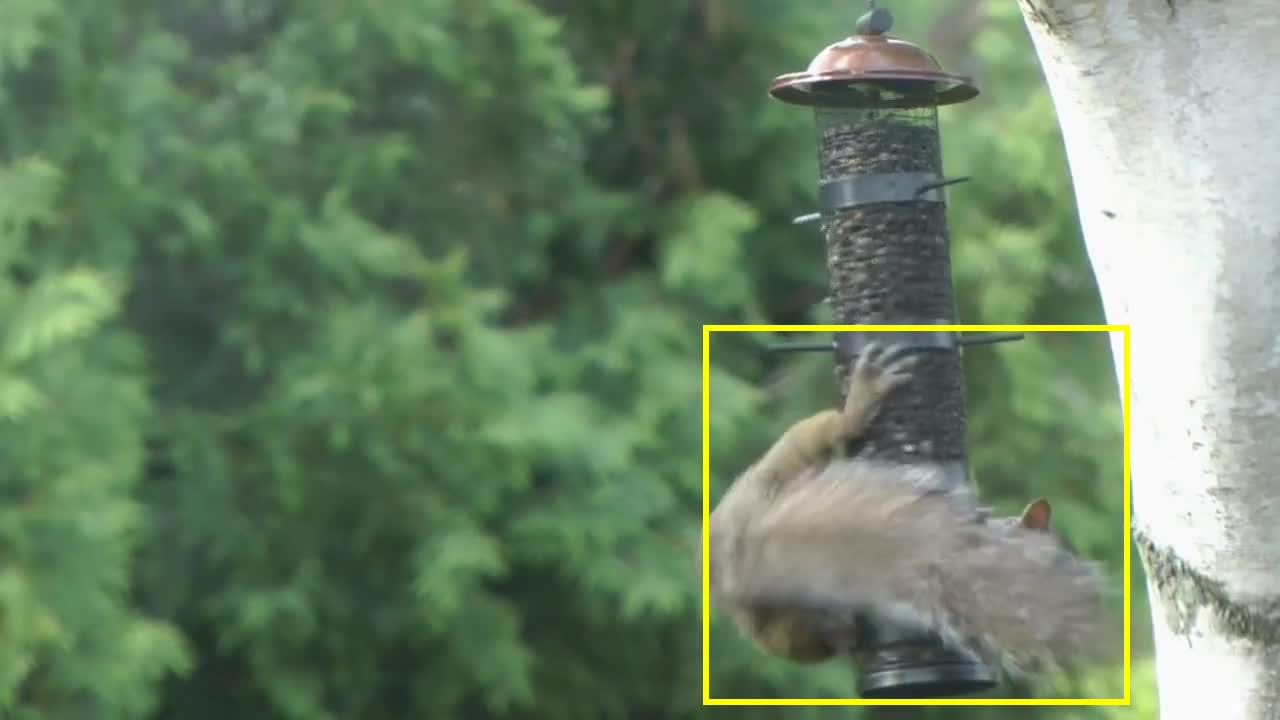}\\
				\includegraphics[trim = 150mm 0mm 90mm 100mm, clip, width=0.25\linewidth]{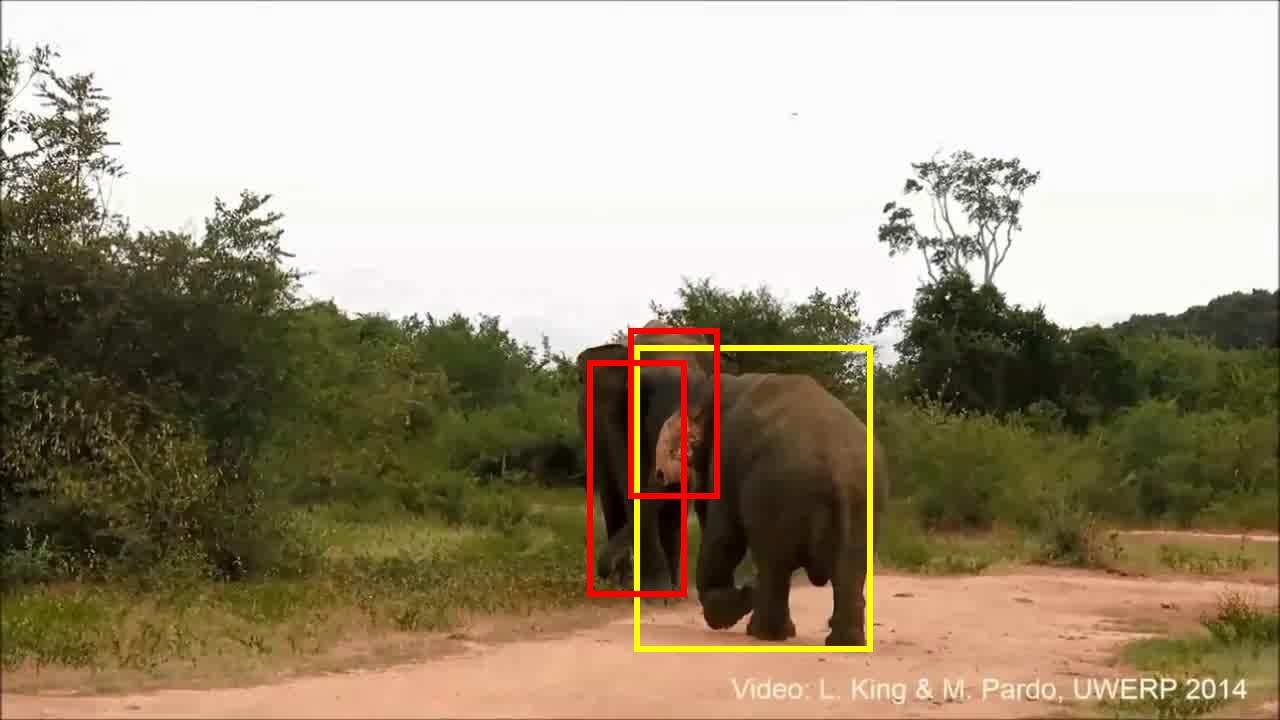}
				&
				\hspace{-0.03cm}\includegraphics[trim = 150mm 0mm 90mm 100mm, clip, width=0.25\linewidth]{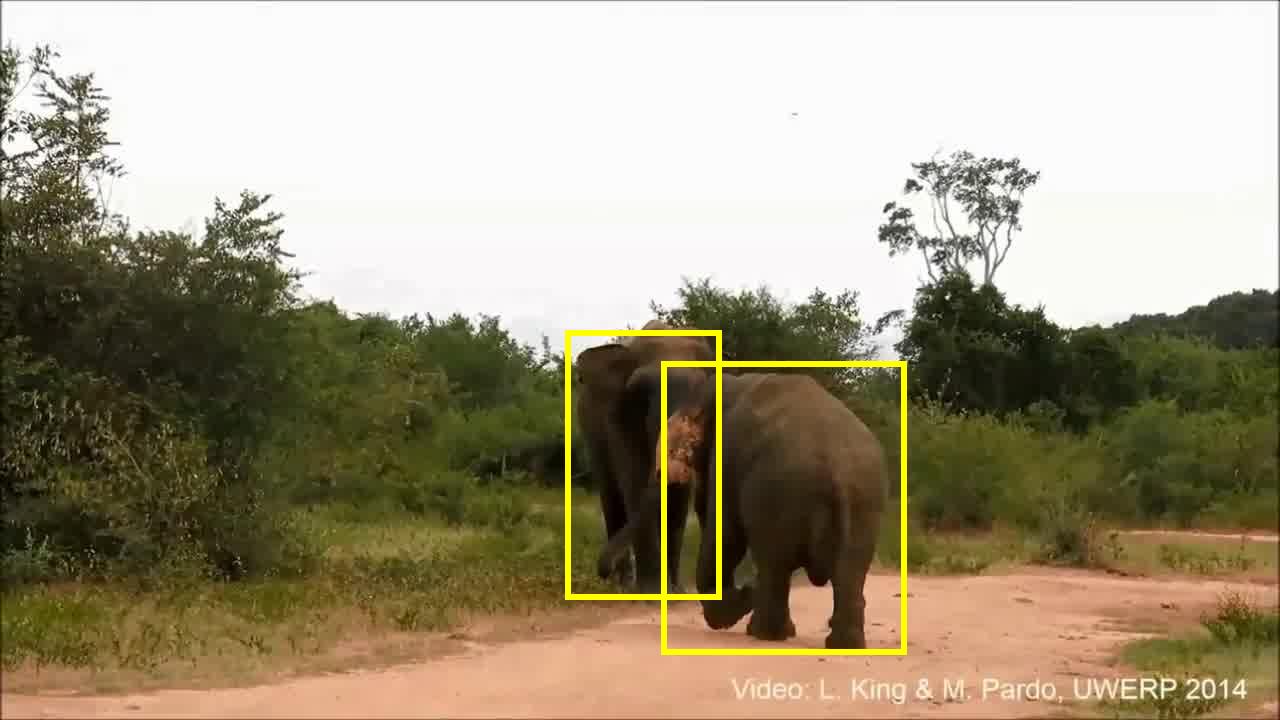}
				&
				\hspace{0.14cm}\includegraphics[trim = 30mm 0mm 0mm 20mm, clip, width=0.25\linewidth]{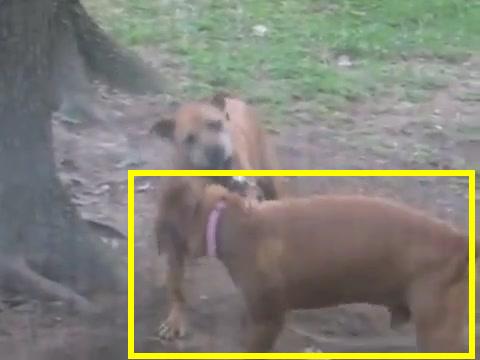}
				&
				\hspace{-0.03cm}\includegraphics[trim = 30mm 0mm 0mm 20mm, clip, width=0.25\linewidth]{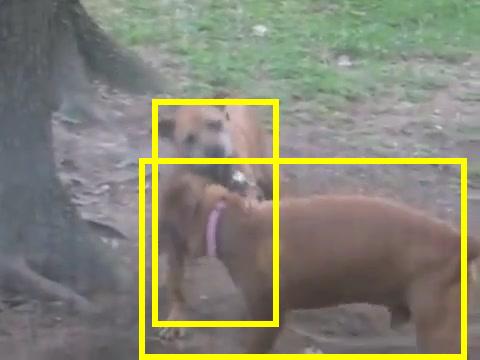}\\
				\includegraphics[trim = 100mm 20mm 50mm 50mm, clip, width=0.25\linewidth]{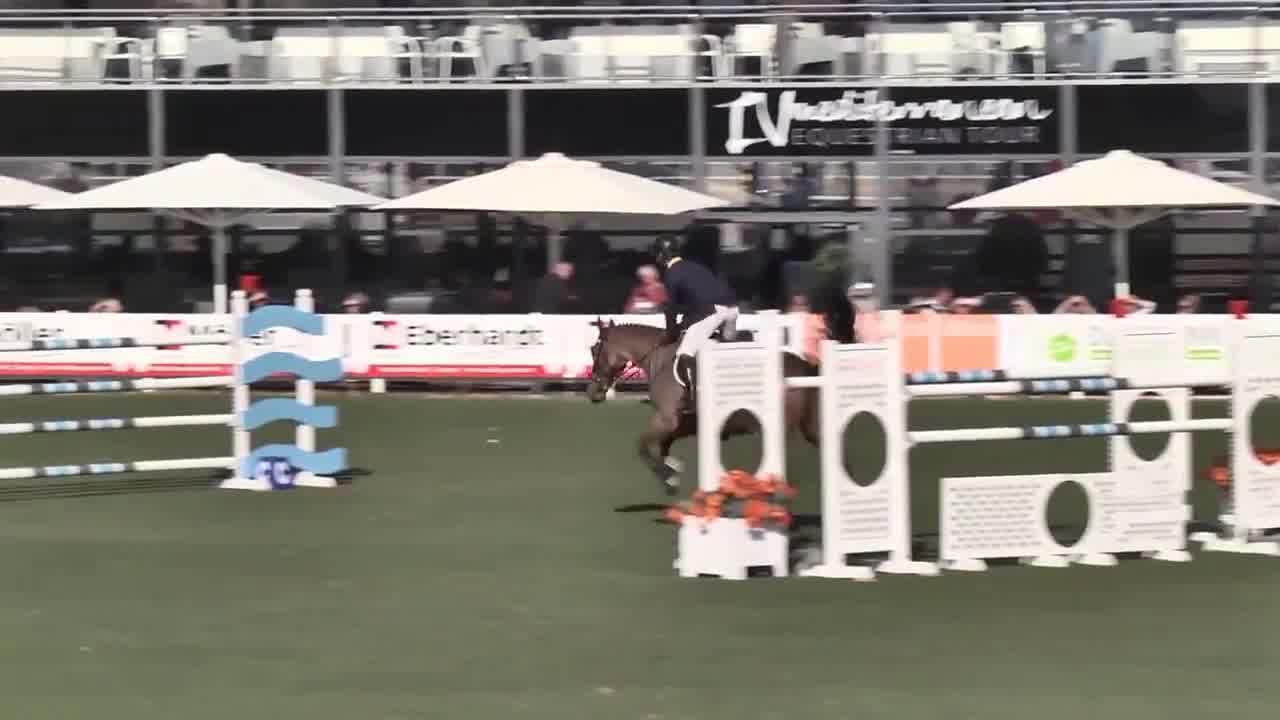}
				&
				\hspace{-0.03cm}\includegraphics[trim = 100mm 20mm 50mm 50mm, clip, width=0.25\linewidth]{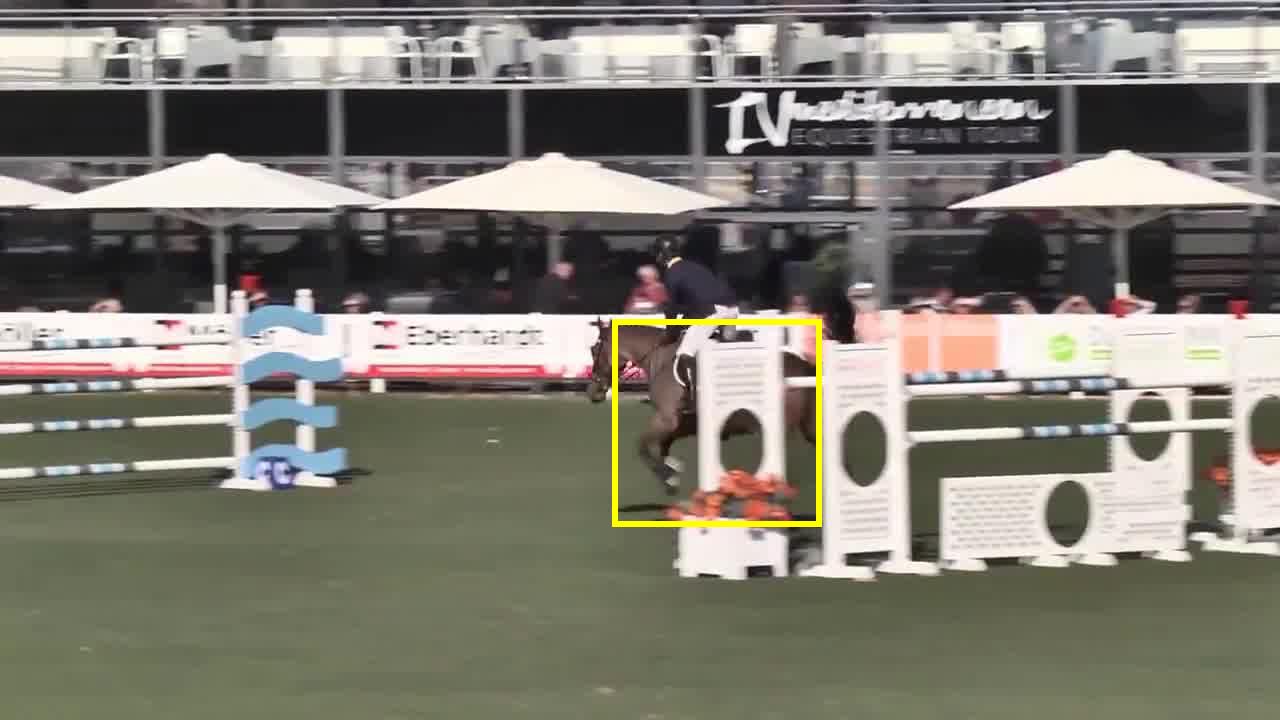}
				&
				\hspace{0.14cm}\includegraphics[trim = 30mm 0mm 0mm 0mm, clip, width=0.25\linewidth]{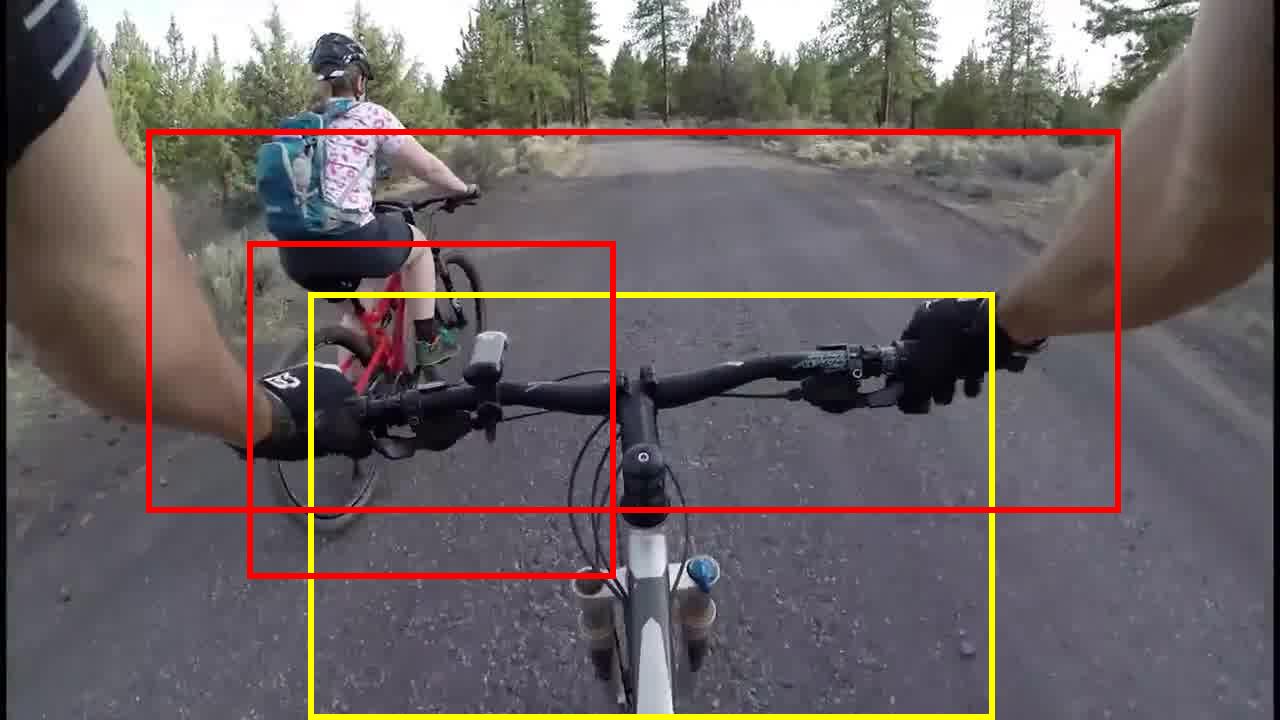}
				&
				\hspace{-0.03cm}\includegraphics[trim = 30mm 0mm 0mm 0mm, clip, width=0.25\linewidth]{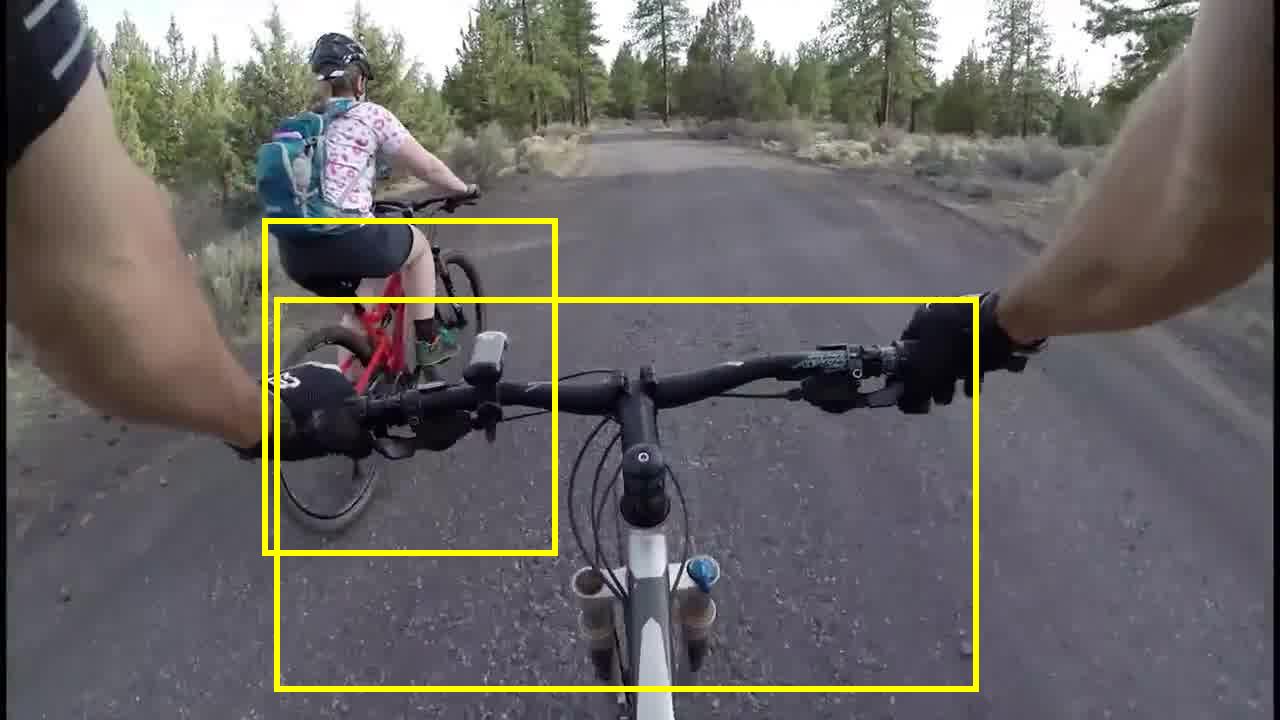}\\
				\includegraphics[trim = 45mm 0mm 0mm 0mm, clip, width=0.25\linewidth]{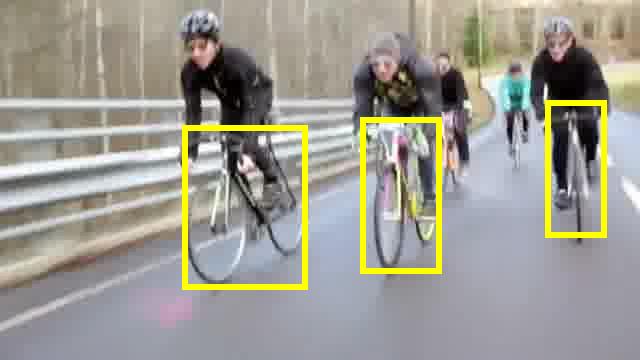}
				&
				\hspace{-0.03cm}\includegraphics[trim = 45mm 0mm 0mm 0mm, clip, width=0.25\linewidth]{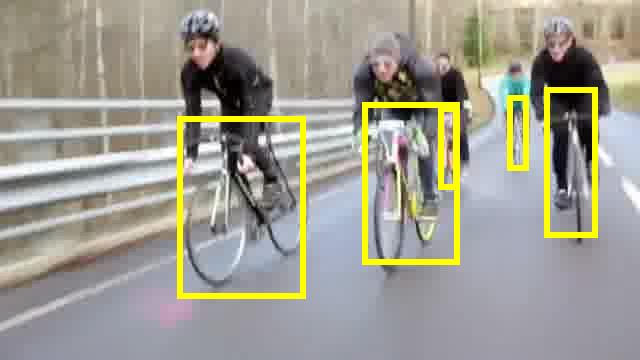}
				&
				\hspace{0.14cm}\includegraphics[trim = 60mm 30mm 70mm 0mm, clip, width=0.25\linewidth]{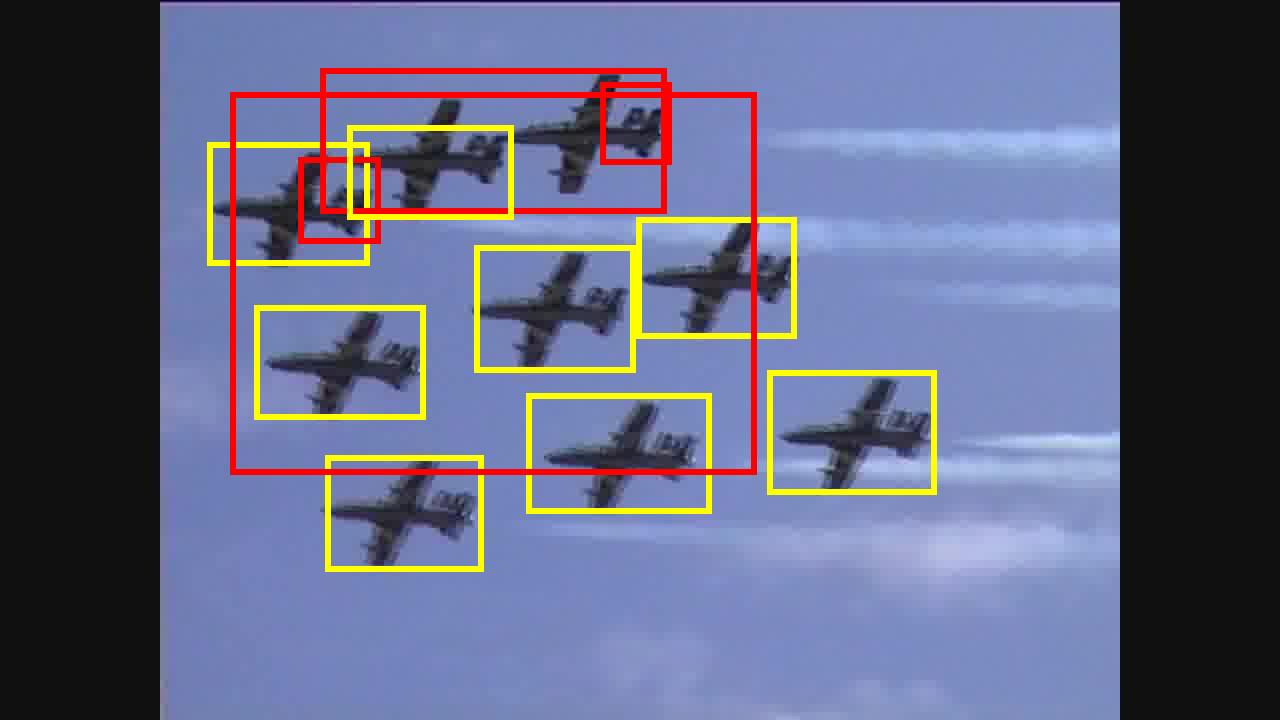}
				&
				\hspace{-0.03cm}\includegraphics[trim = 60mm 30mm 70mm 0mm, clip, width=0.25\linewidth]{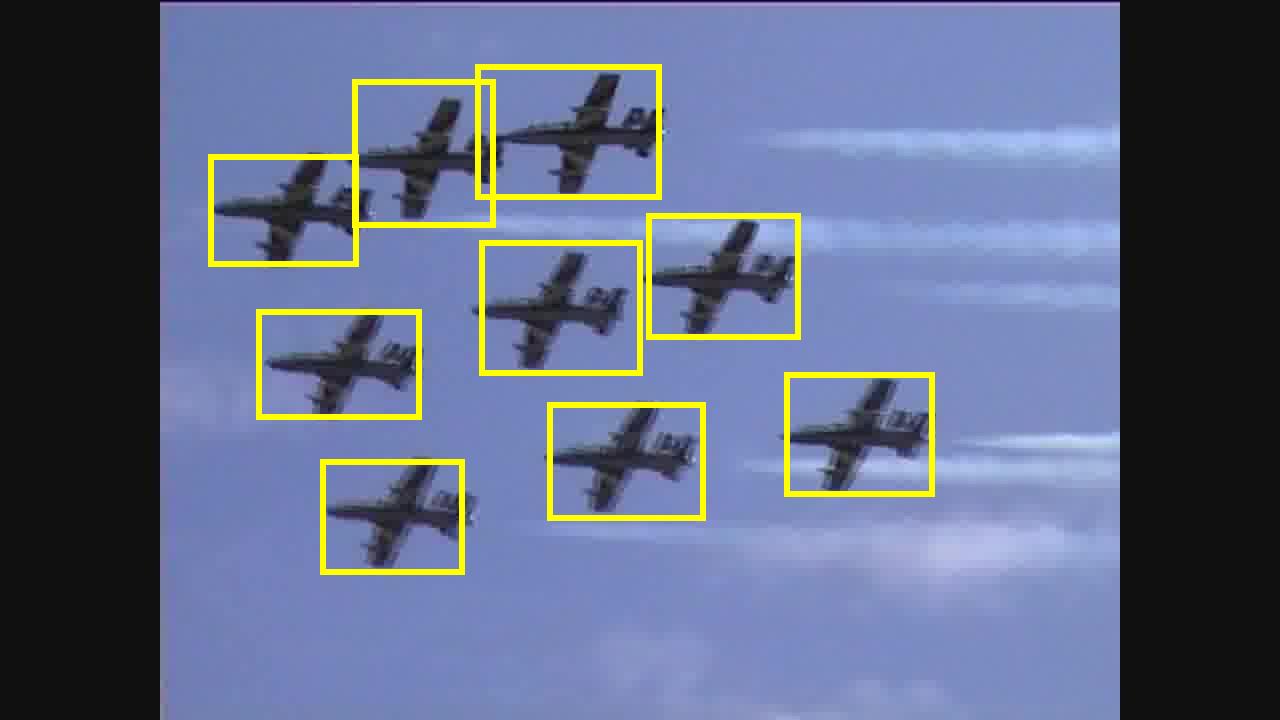}
				
			\end{tabular}
		\end{center}
		\caption{Qualitative results for the Tube-CNN and Box-CNN models for the ImageNet VID dataset. We use the same structure as one in Figure~\ref{fig:supmat_qual_res_casa}.}
		\label{fig:supmat_qual_VID}
	\end{figure*}

	\paragraph{Detection results on ImageNet VID and YTO.}
	We next evaluate Tube-CNN for the more general case of multi-class object detection.
	Table~\ref{tbl:vid_challenge_exp} presents per-class results for the validation set of the ImageNet VID benchmark.
	Similar to experiments on head detection in HollywoodHeads dataset, we observe the consistent improvement of the proposed Tube-CNN over Box-CNN for $28$ out of $30$ object categories.
	We, hence, conclude that our method generalizes well to a large number of object classes.
	The VID column of Table~\ref{tbl:des_performance_HH_Casa} compares results of Tube-CNN to baselines on the ImageNet VID validation set.
	As in the case of head detection, our method improves all tested baselines.
	In Figure~\ref{fig:supmat_qual_VID} we visualize some qualitative results comparing Box-CNN and Tube-CNN.
	
	On YTO we evaluate results using the standard Correct Localization (CorLoc) measure used for this dataset in~\cite{prest2012learning,kang2016object,kang2017object,galteri2017spatio}.
	Given an object class, CorLoc is defined as the percentage of positive images correctly localized according to the PASCAL criterion.
	Table~\ref{tbl:corloc_yto} compares CorLoc results of the Tube-CNN method with the best results reported in the literature.
	Our method provides best performance on YTO even if it was not trained nor fine-tuned on this dataset.
	
	\subsection{Ablation study}
	\label{sec:discuss}
	In this section, we analyze design choices and parameters of our method.
	We also investigate and demonstrate advantages of TPN in terms of computational complexity.
	
	\paragraph{Base architectures and hard negative mining.} 
	Box-CNN and Tube-CNN methods depend on the training schemes, base network architectures as well as the type and the amount of used tube proposals.
	Here we analyze the detection performance by varying each of these factors.
	Table~\ref{tbl:des_performance_HH_detailed} reports performance of our models on HollywoodHeads and ImageNet VID datasets under different settings.
	
	With respect to the training scheme, we achieve consistent improvement with hard negative mining (HN) for all models.
	Regarding the choice of the base network, ResNet architecture (RN) provides significant improvements compared to the more shallow CaffeNet (CN) network.
	Our strongest models use Resnet-101 base network.
	For the ImageNet VID dataset we only report results of models based on Resnet-101.
	
	We train and test Tube-CNN models using tube proposals generated by our TPN model and by tracking box proposals (i.e. SS-Track).
	In Table~\ref{tbl:des_performance_HH_detailed}, we denote these two Tube-CNN models as "Tube-CNN + SS-Track" and "Tube-CNN + TPN" respectively.
	In both cases, Tube-CNN models outperform Box-CNN and other baselines.
	The main advantage of TPN is in test time, which grows linearly with the number of proposals.
	Using a smaller set of proposals, Tube-CNN + TPN is able to achieve similar (or even better) results compared to Tube-CNN + SS-Track.
	This together with the analysis of recall in Section~\ref{sec:recall_ana} confirms the advantage the proposed TPN scheme.
	
	Tube-CNN performs slightly better on SS-Track proposals compared to TPN.
	We believe that Tube-CNN + TPN detection results could be further improved if using deeper networks for the TPN model.
	
	\begin{table}[t]
		\begin{center}
			\hspace*{-0.2cm}
			\tabcolsep=0.12cm
			\begin{tabular}{@{}c||c|cccc@{}}
				Model & \parbox{0.8cm}{\centering Num.\\Prop\vspace{0.1cm}} & CN & CN+HN & RN & RN+HN\\
				\hline
				\hline
				Box-CNN & $\sim$2K & 71.2 & 76.4 & 75.5 & 82.4 \\
				Box-CNN + track & $\sim$2K & 73.1  & 76.9 & 78.0  & 84.7 \\
				\hline
				Tube-CNN + SS-Track & $\sim$6K & \textbf{76.9} & 78.3 & \textbf{83.3} & \textbf{86.8} \\\hline
				\multirow{3}{*}{Tube-CNN + TPN} & 100 & 74.9 & 78.6 & 81.7 & 83.9 \\
				& 300 & 75.8 & 78.7 & 82.3 & 85.0 \\
				& 2K & 73.7 & \textbf{79.1}  & 82.7 & 86.2\\
			\end{tabular}
			\mbox{}
			\subcaption{HollywoodHeads}
			\mbox{}\vspace{.0cm}\\
			\begin{tabular}{@{}c||c|cc@{}}
				Model & \parbox{0.8cm}{\centering Num.\\Prop\vspace{0.1cm}} & RN & RN+HN \\
				\hline
				\hline
				Box-CNN & $\sim$2K & 66.1 & 68.7 \\
				Box-CNN + track & $\sim$2K & 67.2  & 68.9 \\
				\hline
				Tube-CNN + SS-Track & $\sim$17K & - & \textbf{72.7} \\\hline
				\multirow{3}{*}{Tube-CNN + TPN} & 300 & 69.5  & 70.6 \\
				& 1K & 70 & 70.9 \\
				& 2K & 69.8 & 70.9
			\end{tabular}
			\mbox{}
			\subcaption{ImageNet VID}
		\end{center}
		\caption{Detection performance mAP ($\%$) on HollywoodHeads and ImageNet VID dataset. CN and RN and HN stands for CaffeNet, Resnet and hard negative mining respectively.
		}
		\label{tbl:des_performance_HH_detailed}
	\end{table}

	\paragraph{Effect of the tube length.}
	We restrict our method to tubes with linear motion to enable a tractable approach for generating tube proposals.
	We believe linear tubes might be sufficient to represent short time intervals, which is the main target in this work.
	In Table~\ref{tbl:tube_length_exp} we report results comparing performance of Tube-CNN on the HollywoodHeads dataset for different values of tube length $T$ (all models are based on CaffeNet without hard negative mining). We observe the best performance obtained for $T=10$.
	
	\begin{table}[t]
		\begin{center}
			\vspace{-0cm}
			\begin{tabular}{@{}c||ccccc@{}}
				Tube length & T=1 & T=5 & T=10 & T=15 & T=20\\
				\hline
				\hline
				AP (\%) & 71.2 & 76.6 & \textbf{76.9} 
				& 73.4 & 70.1\\[-0.6cm]
			\end{tabular}
		\end{center}
		\caption{
			Tube-CNN performance for different values of T.}
		\label{tbl:tube_length_exp}
	\end{table}
	
	\paragraph{Advantage of feature-level aggregation.}
	In this experiment, we show the advantages of early feature aggregation with TOI pooling and learning features for tube classification.
	We compare our Tube-CNN framework with baselines of temporal aggregation of single-frame image features.
	In particular, we use our best CaffeNet-based single frame Box-CNN detector to extract FC7 features on all the frames of training tubes and train a linear SVM classifier based on the temporally aggregated features.
	SVM parameters for each setup are chosen with 5-fold cross-validation.
	In Table~\ref{tbl:agg_boxcnn} we report results for max and average aggregation baselines.
	The improvement of Tube-CNN over baselines confirms the advantage of jointly learning features and their temporal aggregation as proposed in this work.
	
	\begin{table}[t]
		\begin{center}
			\begin{tabular}{@{}c|cc|c@{}}
				Box-CNN & fc7 max & fc7 avg & Tube-CNN\\
				\hline
				\hline
				76.4 & 77.0 & 77.3 & {\bf 79.1}\\[-0.6cm]
			\end{tabular}
		\end{center}
		\caption{
			Aggregation of Box-CNN and Tube-CNN.\vspace{-0.1cm}}
		\label{tbl:agg_boxcnn}
	\end{table}

	\paragraph{Running time.}
	Table~\ref{tbl:TPN_speed} compares running times of object detectors (in frames per second) for Tube-CNN on SS-Track and TPN proposals.
	To obtain results for SS-Track, we accumulate the time required for per-frame Selective Search proposals, KLT point tracking and Tube-CNN evaluation.
	Results for TPN are obtained by summing times for the evaluation of  the TPN network, applying tube-NMS and evaluating Tube-CNN.
	When using ResNet-based Tube-CNN object detectors, our method with 100 TPN is about $28$ times faster compared to Tube-CNN + SS-Track.
	The improvements of running time come from the efficient generation of TPN proposals and from the reduced number of proposals required for the Tube-CNN evaluation.
	
	\begin{table}[t]
		\begin{center}
			\hspace*{-0.5cm}
			\tabcolsep=0.12cm
			\begin{tabular}{@{}c||c|c|cccc@{}}
				Method&SS-Track&RPN-Track&\multicolumn{4}{c}{TPN}\\
				\hline
				\hline
				Num.TP&$\sim$6K&$\sim$1.2K&100&300&500&1K\\
				\hline
				CN&0.3&2.3&\textbf{4.3}&3.7&3.0&2.3\\
				RN&0.06&0.7&\textbf{1.7}&1.1&1.0&0.7\\[-0.6cm]
			\end{tabular}
		\end{center}
		\caption{Running times (frames per second) for Tube-CNN object detectors using SS-Track proposals and different numbers of TPN proposals.
		}
		\label{tbl:TPN_speed}
	\end{table}
	
	\section{Conclusion}
	\label{sec:conclusion}
	We have addressed the task of object detection in video.
	To this end, we proposed two CNN models for generating and classifying object tube proposals respectively.
	Our tube classification model achieves state-of-the-art results on all four tested datasets for object detection in video.
	The unified framework of the two models is both accurate and computationally efficient at the running time.
	Our method particularly improves object detection in difficult situations with dynamic occlusions.
	We therefore believe it could be highly valuable when used as input for object tracking and other challenging scenes.
	
	\paragraph{Acknowledgements.} This work was supported in part by the ERC grant ACTIVIA and the MSR-Inria joint lab.
	
	\appendix
	\section{Training Tube-CNN}
	\label{sec:train_tube_cnn}
	\paragraph{Training procedure.}
	We optimize Tube-CNN with the stochastic gradient descent (SGD) algorithm using momentum $0.9$ and weight decay $0.0005$ on mini-batches.
	Mini-batch sampling is done hierarchically.
	We first sample $4$ chunks of consecutive frames.
	Then, on each sampled chunk, we sample a set of $64$ tube proposals.
	The training batch therefore contains $4 \times 64$ tube proposals in total. 
	Positive samples contribute at most $25\%$ of the training batch.
	We initialize the learning rate at $0.001$, and decrease it several times by a factor of $10$ after the validation error reaches saturation.
	
	\paragraph{Network architectures.}
	We use two ImageNet pre-trained networks to initialize Tube-CNN model: CaffeNet~\cite{girshick2015fast} and Resnet-101~\cite{He2015resnet}.
	Our training settings for these two architectures are descrived below.
	
	\paragraph{CaffeNet.} 
	Tube-CNN initialized from CaffeNet has a CNN feature extractor block composed of the first $5$ convolutional layers.
	The last fully-connected (FC) layers are used for tube classification and tube regression.
	TOI pooling used for tube classification, and ROI pooling used for tube regression are applied to the feature volume created by stacking $conv5$ feature maps.
	
	As described in Section~\ref{sec:tube_cnn}, our TOI pooling consists of: spatial max-pooling and temporal max-pooling.
	Similarly to~\cite{girshick2015fast}, the spatial max-pooling produces fixed-size $6\times 6$ output features.
	On the other hand, temporal max-pooling collapses all values along temporal dimension into one max value.
	Therefore, for every input tube proposal, TOI pooling layer outputs a $256$-channels feature map with a fixed spatio-temporal extent of $6\times 6\times 1$.
	Each feature map is flattened into a $9126$-dimentional feature vector, which is then passed through FC layers to compute the classification score.
	
	We use the same spatial hyper-parameters of $6\times 6$ for ROI pooling.
	Given an input tube proposal, the ROI pooling layer outputs two $9126$-dimensional feature vectors corresponding to the two ends of the tube proposal.
	We pass the two feature vectors through FC layers to produce regression parameters for the beginning and the end of the tube.
	
	\paragraph{Resnet-101.}
	TOI pooling and ROI pooling layers are inserted between the third and the fourth blocks of the network, i.e. \textit{conv\_4x} and \textit{conv\_5x} of Table 1 in~\cite{He2015resnet}.
	In detail, $91$ convolutional layers before and including \textit{conv\_4x} belong to CNN feature extractor block of Tube-CNN.
	All layers after and including \textit{conv\_5x} construct tube classification and tube regression blocks.
	
	The spatial extent of both TOI and ROI pooling layers has a fixed-size of $7\times 7$.
	The temporal max-pooling is done similarly as for CaffeNet.
	Without being flattened, pooled feature maps are passed into $conv\_5x$ followed by an average pooling layer.
	In our setting, output size of $conv\_5x$ is different from the one of the original network.
	We adapt to this change by adjusting kernel size of the average pooling layer to $4\times 4$.
	The last FC layers of tube classification and tube regression blocks take the average pooling output to produce final object score and position.
	
	\section{Training TPN}
	\label{sec:train_tpn}
	
	\paragraph{Training batch.} 
	TPN model is trained on mini-batches of tube anchors.
	Similar to the Tube-CNN, mini-batches are hierarchically sampled.
	The training batch contains $4 \times 128$ tube anchors in total. 
	Different to Tube-CNN, TPN supervision is class-agnostic, i.e. each tube anchor is labeled as either an object (positive) or background (negative).
	In our setting, at most $50\%$ of the training batch are positive samples.
	
	We optimize the TPN model with an SGD algorithm.
	The model is only initialized with the CaffeNet~\cite{girshick2015fast} based network.
	The learning rate of SGD is set to~$0.001$, momentum~-- to~$0.9$, weight decay~-- to~$0.0005$.
	We have reduced the learning rate by a factor of $10$ after saturation of the validation error.
	
	\paragraph{Tube anchor design.}
	Our tube models are trained on two datasets: HollywoodHeads~\cite{vu2015context} and ImageNet VID~\cite{russakovsky2015videoImageNet}.
	While Tube-CNN hyper-parameters are the same on the two datasets, the ones of TPN are adapted to each dataset, i.e. different tube anchors are used.
	
	As mentioned in Section~\ref{sec:tpn}, we use tube anchors with spatial location fixed, but with varying scales and aspect ratios.
	On both datasets, tube anchors have $6$ fixed scales.
	On HollywoodHeads dataset, because human heads are mostly square, we impose only one aspect ratio 1:1 to all tube anchors.
	On ImageNet VID dataset, as objects have more diverse shapes, the TPN achieves best object recall with $5$ aspect ratios 1:4, 1:2, 1:1, 2:1 and 4:1.
	
	\section{Hard negative mining}
	\label{sec:hardnegdetails}
	As discussed in Section 5.2 and Section 5.3, our best Box-CNN and Tube-CNN models are trained with a few passes of hard negative mining.
	We here describe the procedure in an abstract manner.
	The term \textit{proposal} stands for box proposal in case of Box-CNN, and tube proposal in case of Tube-CNN.
	The overlap ratio between proposal and ground-truth is defined separately for each model.
	
	An iteration of hard negative mining consists of three steps: (1) a forward pass on a subset of input proposals to find hard negatives; (2) composition of several training batches with hard negatives and random positives; (3) training the network on the constructed batches.
	We define hard negatives as high-scoring false positives with no overlap to any ground-truth.
	The hard negative training batch contains $25\%$ positives and $50\%$ hard negatives at most.
	The rest of the batch is composed of negatives having overlap ratio to ground-truth in the range $[0.1,0.5)$.
	
	{\small
		\bibliographystyle{ieee}
		\bibliography{egbib}
	}
\end{document}